\documentclass[preprint,3p,times,12pt]{elsarticle}
\usepackage{amsmath,amsfonts}
\usepackage{algorithmic}
\usepackage{array}
\usepackage[caption=false,font=normalsize,labelfont=sf,textfont=sf]{subfig}
\usepackage{textcomp}
\usepackage{stfloats}
\usepackage{url}
\usepackage{verbatim}
\usepackage{lineno}
\usepackage{setspace}
\usepackage{graphicx}
\usepackage{pdflscape}

\def\BibTeX{{\rm B\kern-.05em{\sc i\kern-.025em b}\kern-.08em
    T\kern-.1667em\lower.7ex\hbox{E}\kern-.125emX}}
\usepackage{balance}

\usepackage{booktabs}
\usepackage{epsfig}
\usepackage{multirow}
\usepackage{makecell}
\usepackage{pgfgantt}
\usepackage{pdflscape}
\usepackage{enumitem}
\usepackage{rotating}
\usepackage{afterpage}

\begin{document}
\title{Integrating Semi-Supervised and Active Learning for Semantic Segmentation}
\author{
Wanli Ma\textsuperscript{1,2}, 
Oktay Karakuş\textsuperscript{2}, 
Paul L. Rosin\textsuperscript{2} \\
\textsuperscript{1}Department of Engineering, University of Cambridge, Cambridge CB2 1TN, U.K.\\
\textsuperscript{2}School of Computer Science and Informatics, Cardiff University, Cardiff CF24 4AG, U.K. 
}

\markboth{Journal of \LaTeX\ Class Files,~Vol.~18, No.~9, September~2020}%
{How to Use the IEEEtran \LaTeX \ Templates}

\begin{abstract}
Pixel-level annotation for image segmentation tasks is both time-consuming and expensive, particularly in the context of remote sensing. To mitigate this challenge, semi-supervised learning and active learning offer effective solutions, although they are typically employed independently. In this paper, we propose a novel hybrid learning framework that integrates active learning with an enhanced semi-supervised learning strategy, tailored to reduce annotation costs and improve semantic segmentation performance, particularly in the domain of remote sensing, where dense pixel-level annotation is expensive and labour-intensive. Our method leverages both the labelled samples selected via active learning and the unlabelled data excluded from selection by incorporating a pseudo-label auto-refinement (PLAR) mechanism. This module identifies potentially inaccurate pseudo-labels and automatically refines them based on the cluster assumption that similar features in high-density regions of the feature space correspond to the same class. Importantly, manual annotation is limited to only the most uncertain regions, while less ambiguous areas are refined without consuming labelling budget. We evaluate our proposed framework on multiple benchmark datasets covering both natural and remote sensing imagery. The results demonstrate that our method consistently outperforms state-of-the-art baselines in image segmentation tasks, confirming its effectiveness and generalisation capability across different domains. %The code is available at https://github.com/WANLIMA-CARDIFF/PLAR.
% In this paper, we propose a novel active learning approach integrated with an improved semi-supervised learning framework to reduce the cost of manual annotation and enhance model performance. Our proposed approach effectively leverages both the labelled data selected through active learning and the unlabelled data excluded from the selection process. 
% % By assigning pseudo-labels, the unlabelled data is incorporated into the training process, allowing the model to learn from it despite the absence of true labels. 
% The proposed active learning approach pinpoints areas where the pseudo-labels are likely to be inaccurate. Then, an automatic and efficient pseudo-label auto-refinement (PLAR) module is proposed to correct pixels with potentially erroneous pseudo-labels by comparing their feature representations with those of labelled regions. This approach operates without increasing the labelling budget and is based on the cluster assumption, which states that pixels belonging to the same class should exhibit similar representations in feature space. Furthermore, manual labelling is only applied to the most difficult and uncertain areas in unlabelled data, where insufficient information prevents the PLAR module from making a decision. We assessed the proposed hybrid semi-supervised active learning framework across multiple benchmark datasets spanning both natural and remote sensing domains. In both cases, it outperformed state-of-the-art methods in the semantic segmentation task.

\end{abstract}


\maketitle

% \linenumbers
\onehalfspacing

\section{Introduction} 
\label{sec:intro}

Image segmentation plays an important role in earth observation applications, including land cover classification, urban monitoring, and crop type mapping \cite{ma2022amm, wang2022unetformer, pott2021satellite}. Deep learning networks have shown great success in this domain, yet their performance is critically dependent on the availability of large-scale, high-quality annotated datasets. This requirement is especially burdensome in remote sensing, where pixel-level annotation of satellite or aerial imagery is both time-consuming and costly. Compared to natural images, remote sensing imagery introduces further complexity due to spectral variability, large spatial scales, and the need for expert knowledge in labelling \cite{yue2022optical}. Consequently, strategies that reduce annotation burden while maintaining segmentation quality are vital for advancing automated interpretation of remote sensing data.
 
Semi-supervised learning (SSL) is one such approach, aiming to leverage a small amount of labelled data in combination with abundant unlabelled data. SSL methods help reduce the dependence on large annotated corpora, making them highly appealing for remote sensing applications where exhaustive labelling is impractical. Among these, Teacher-Student frameworks and Cross Pseudo Supervision (CPS) have shown promise in fusing labelled and unlabelled information to boost performance. However, limitations remain: Teacher-Student frameworks often suffer from over-reliance on the teacher, while CPS-based methods can collapse as model weights converge during training, diminishing the benefits of dual-network interaction. To address these challenges, we propose a novel hybrid framework, \textit{Teacher-Student-Friend} (TSF), which incorporates an auxiliary “friend” model into the learning loop, unifying the strengths of both paradigms and offering greater resilience during training.
 
In parallel, active learning (AL) has emerged as a principled way to prioritise the most informative samples for annotation. While SSL and AL are often treated independently, their combination is particularly advantageous for remote sensing. In this work, we advocate a unified strategy that blends SSL and AL: while SSL exploits the bulk of unlabelled data, AL strategically selects samples for labelling based on their uncertainty or diversity. Notably, existing AL approaches often discard unlabelled samples not selected for annotation. We argue this does not fully leverage data; those unselected samples still contain exploitable structure that can assist learning when pseudo-labelled properly.
 
\begin{figure}
    \centering
    \includegraphics[width=0.9\linewidth]{ 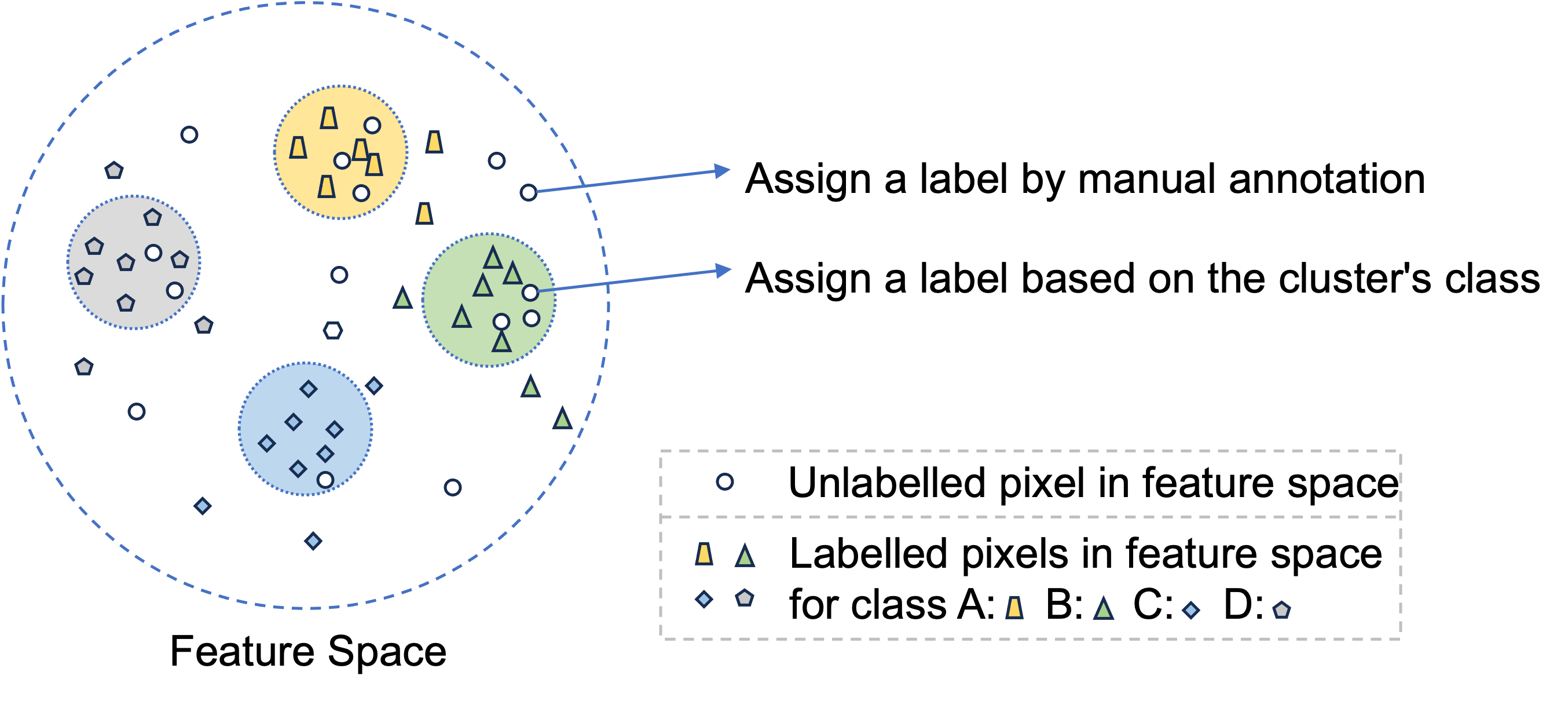}
    \caption{Pseudo-label auto-refinement based on similarity in feature space. }
    \label{fig:Illustration_igure}
\end{figure}

To bridge this gap, we introduce an active learning strategy designed to refine pseudo-labels. It identifies label inconsistencies by examining feature similarity between labelled and unlabelled regions within the same image. Grounded in the assumption that spatially or spectrally similar regions in remote sensing images are likely to belong to the same class \cite{4787647}, we propose an efficient pseudo-label auto-refinement mechanism. This mechanism operates in feature space and enables automatic correction of unlabelled pixel predictions by referencing the surrounding context of known labels, as shown in Figure \ref{fig:Illustration_igure}. It provides a form of soft supervision without additional manual annotation, further conserving the labelling budget.

The key contributions of this work are summarised below:
\begin{itemize}
\setlength\itemsep{1pt}
    \item We propose a novel integration of active learning and semi-supervised learning for semantic segmentation, validated across both natural and remote sensing imagery.
    \item We develop a robust hybrid framework, Teacher-Student-Friend (TSF), that combines the strengths of Student-Teacher models and Cross Pseudo Supervision to improve learning from limited labels.
    \item We introduce a pseudo-label auto-refinement strategy using feature similarity within images, enabling the correction of noisy pseudo-labels in unlabelled regions with minimal cost.
    \item We present a dedicated module for identifying high-value samples in the active learning process, using both prediction uncertainty and multi-level feature representations.
\end{itemize}

% (1)  (2)  (3)  (4) 

% \begin{enumerate}[topsep=0pt,itemsep=0pt]
%     \item We propose a novel semi-supervised learning (SSL) framework, called Student-Teacher-Friend (STF), which combines and exploits the capabilities of the student-teacher SSL framework with cross-pseudo supervision (CPS).
%     \item We introduce an effective combination of active learning and semi-supervised learning for semantic segmentation applications that efficiently make use of unlabelled data.
%     \item We propose an automatic labelling strategy for active learning that leverages feature similarity between unlabelled regions and labelled areas within a single image to assign estimated labels to valuable unlabelled areas. By utilizing Euclidean and Mahalanobis distances for this process, our approach ensures reliable estimated labels and significantly lowers the labelling costs associated with active learning.
%     \item We develop a reliable module for active learning procedures that enhances the identification of the most informative unlabelled data for pixel-wise labelling. This module takes into account not only the probability map of the predictions but also examines multi-level feature representations of the unlabelled data samples.
    
% \end{enumerate}

\section{Related Work}
\subsection{Semi-supervised semantic segmentation} 
In the field of semi-supervised learning, consistency regularisation is widely explored and primarily aims to enforce neural networks to produce consistent predictions by incorporating various types of perturbations. The strategies for the said perturbations can be categorised into \textit{input} \cite{french2019semi, olsson2021classmix, zou2020pseudoseg}, \textit{feature} \cite{ouali2020semi}, network \cite{chen2021semi} and \textit{hybrid}. 

Specifically, the Mean Teacher (MT) \cite{tarvainen2017mean} and Cross Pseudo Supervision (CPS)~\cite{chen2021semi} are two well-known fundamental consistency learning frameworks. The former, MT, employs a teacher model whose parameters are the exponential moving average (EMA) of a student model's parameters. A few variations of this approach can be found in the literature (See e.g., \cite{xu2022semi, hu2021semi}). AEL \cite{hu2021semi} introduces an adaptive CutMix and sampling strategy for unlabelled data, designed to enhance learning for underperforming classes based on the evaluation of the current model. U2PL \cite{wang2022semi_u2pl} leverages not only highly confident predictions for unlabelled data as pseudo-labels for training the networks but also utilises ambiguous predictions as negative samples, helping the network learn by contrasting them with the corresponding positive samples. Dual Student \cite{ke2019dual} argued that the EMA-based Teacher-Student leads to a performance bottleneck due to the tight coupling between the two roles, with this interdependence intensifying as training progresses. 

On the other hand, the latter, CPS, leverages unlabelled data differently from the Teacher-Student structures by enforcing consistency between two models of the same architecture with different initialisation. Some other hybrid SSL approaches include ST++ \cite{yang2022st++}, ReCo \cite{liu2021bootstrapping}, and S4AL+ \cite{rangnekar2022semantic}, integrating self-training, contrastive, or representation learning. Recently, PEFAT, a semi-supervised framework, was proposed to leverage pseudo-loss estimation and feature-level adversarial training to improve multi-class and multi-label medical image classification \cite{10203490}. LeFeD was proposed to leverage feature-level discrepancies from dual decoders to improve volumetric medical image segmentation \cite{10619990}. VerSemi, a versatile semi-supervised framework, was proposed to unify multiple SSL tasks with dynamic task prompts and synthetic augmentation to improve medical image segmentation \cite{10945994}. PICK combines pseudo-label-guided masking, reconstruction, and multi-decoder learning to improve medical image segmentation \cite{zeng2025pick}.

It is important to note that the proposed STF framework differs from Dual Student by integrating CPS with the EMA-based Teacher-Student. CPS, introduced after Dual Student, uses pseudo-labels rather than a consistency constraint, effectively expanding the training data. This combination leverages the strengths of both training structures and aims to overcome the performance limitations of using either the Teacher-Student or CPS method alone (See Section \ref{sec:tsf} for more details). %Specifically, in teacher-student methods, the quality of pseudo-labels depends heavily on the performance of the teacher model. However, by combining this approach with CPS, the pseudo-labels used to train the student are sourced from both the teacher and friend models, which have minimal connection to each other. This integration helps reduce the over-reliance on the teacher model’s performance, addressing a key limitation that can impair the efficiency of model training, particularly in situations with high levels of noisy data.

Adjacent Teacher \cite{xia2025adjacent} observed a similar phenomenon in which objects of the same category, or those that are closely related, tend to exhibit a clustered spatial distribution and similar orientation. By leveraging neighbouring information, their approach focuses on object-wise pseudo label correction in SSL; our method operates at the pixel level using the feature with nearby distribution. Additionally, for pseudo labels that cannot be reliably corrected, the proposed method introduces a supplementary strategy involving limited manual annotation to address these challenging instances by active learning.

\subsection{Active learning }
In the literature, active learning mainly focuses on proposing methodologies to measure the importance of each unlabelled sample for training to be manually annotated, thereby aiding the training process of deep learning networks. The literature mostly defines the importance based on either uncertainty \cite{yoo2019learning, xie2020deal, gorriz2017cost, gal2017deep} or diversity \cite{sener2017active, jain2016active, sinha2019variational}. In semantic segmentation, based on the granularity of annotation, active learning can be classified under three categories: image-level, region-level, and pixel-level. 

Image-level active learning approaches use the whole image as a minimal evaluation unit, aiming to select the most informative samples for labelling. Core-set \cite{sener2017active} is a foundational method that selects a small, representative subset of data such that a model trained on this subset performs well on the entire dataset. It does this by minimising a theoretical loss bound, which is equivalent to solving a k-Center problem, effectively ensuring coverage of diverse data points. Yoo and Kweon \cite{yoo2019learning} proposed utilising a loss prediction module to estimate target losses for unlabelled data. These predicted losses can then be used to identify images on which the target model is likely to make incorrect predictions. Similarly, DEAL \cite{xie2020deal} employed a semantic difficulty branch to predict difficulty scores for different semantic areas within an image, and then selects samples with the highest predicted difficulty, emphasising regions that contribute most to improving model performance. While image-level approaches provide simplicity and global coverage, they may overlook fine-grained structures and local ambiguities, which can be critical in tasks such as semantic segmentation.

Region-level approaches evaluate and select informative regions within images for training. ViewAL \cite{siddiqui2020viewal} proposed a viewpoint entropy formulation that integrates prediction uncertainty to identify the most informative regions in images. Casanova et al. \cite{casanova2020reinforced} introduced a reinforcement learning approach for active learning, the Deep Q-network, to select regions based on the predictions and uncertainty of a segmentation model.  
Colling et al. \cite{colling2020metabox+} utilised a meta-regression model to identify informative regions in the training data and estimate annotation costs, aiming to target regions that are both informative and have low annotation costs. S4AL \cite{rangnekar2023semantic} leverages a teacher-student framework to generate pseudo-labels for identifying regions that help differentiate confused classes, and introduces mechanisms to enhance performance on imbalanced label distributions, which were previously unaddressed in active learning for semantic segmentation. Region-level approaches, by focusing on informative subregions, are particularly effective for dense prediction tasks, but typically require more sophisticated models to evaluate and compare local areas.

Inspired by PixelPick \cite{shin2021all}, the proposed method identifies uncertain data at the pixel level and introduces an auto-labelling module to annotate these pixels with estimated labels, followed by manual labelling.

\afterpage{
  \clearpage
  \begin{landscape}
\begin{figure*}[p]
    \centering
    \includegraphics[width=\linewidth]{ 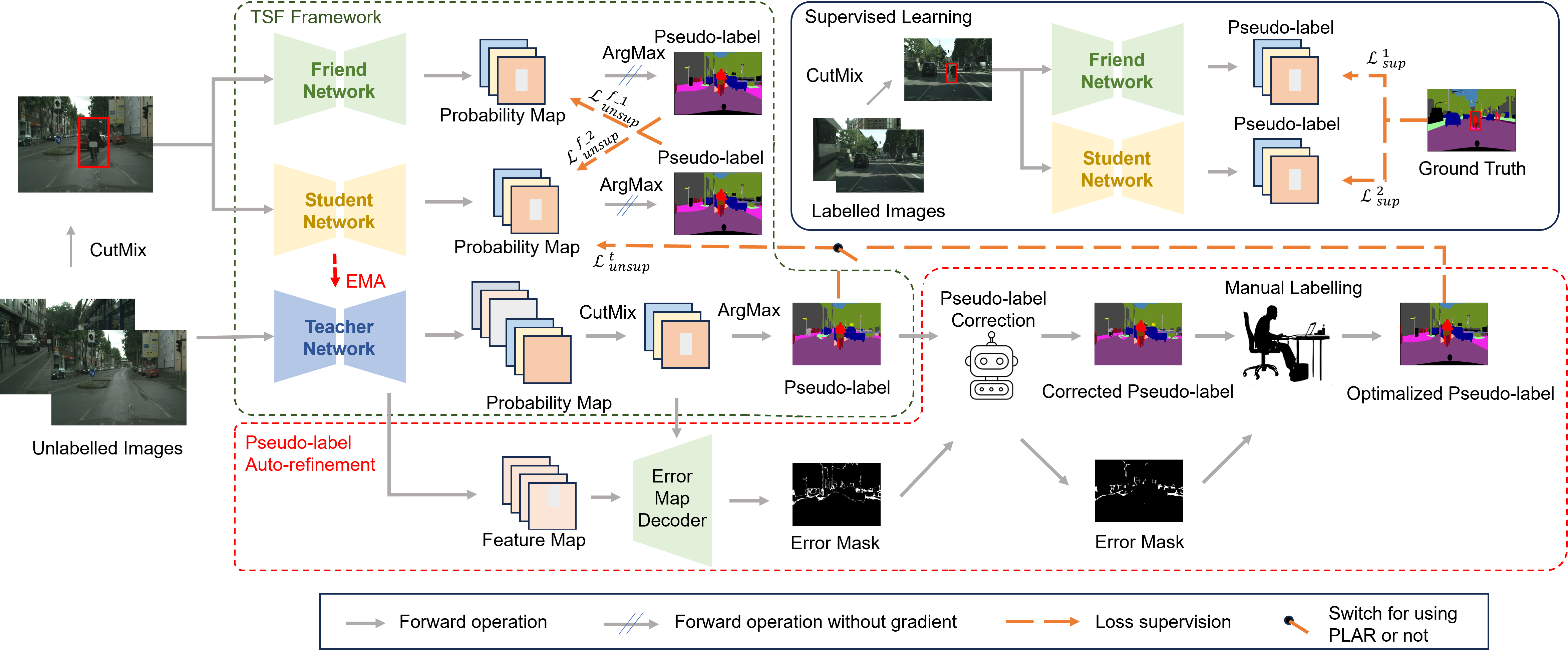}
    \caption{The proposed hybrid active semi-supervised framework integrates the Teacher-Student-Friend (TSF) SSL approach with an active learning strategy, leveraging a novel pseudo-label auto-refinement (PLAR) module to enhance the quality of unlabelled data.} 
    \label{fig:SSL_framework}
\end{figure*}
\end{landscape}
\clearpage
}

\section{The Proposed Method}
The proposed method integrates an improved SSL framework with a novel active learning strategy to enhance the quality of pseudo-labels. For the proposed active learning stage, uncertain regions are identified using two complementary strategies. First, we employ an error map decoder (EMD), which is designed to directly predict areas that are likely to contain errors in the pseudo labels. The EMD is trained using the labelled part of the training data, where the discrepancy between pseudo labels and ground-truth annotations is used to supervise the learning of potential error regions. During the training phase, EMD is learned in a training-and-use manner. Subsequently, in the active learning stage, it is employed in inference mode to generate an error probability map for unlabelled data, highlighting regions with a high likelihood of prediction errors.
Second, we estimate uncertainty based on the prediction confidence of the teacher model. Specifically, the probability outputs of the teacher model for unlabelled data are used to identify low-confidence regions. A predefined threshold is applied to the probability map to select pixels or regions with low prediction confidence, which are considered uncertain. 
Finally, the uncertain regions obtained from the EMD-generated error map and the confidence-based uncertainty map are combined using a logical \textit{`OR'} operation. This fusion strategy enables the framework to capture potential error regions from both perspectives, thereby improving the coverage of uncertain areas for the subsequent active learning process. Importantly, a pseudo-label auto-refinement (PLAR) module to correct these errors, which does not consume manual labelling budget. The remaining errors that the PLAR cannot address are corrected through manual labelling. % of a small subset of regions.

% \subsection{The Proposed Semi-supervised Learning Framework}
\subsection {The Teacher-Student-Friend (TSF) Structure}\label{sec:tsf}

As discussed in the above sections, teacher-student-based frameworks can lead to over-reliance on the teacher's performance, especially with noisy data, while CPS often results in model similarity that limits the benefit of parallel learning. To provide a solution for the problem mentioned above, 
%that will enhance the diversity of training samples for the student model and , 
we propose incorporating an additional ``friend" model into the Teacher-Student framework, called TSF, as shown in Figure \ref{fig:SSL_framework} to ensure the student model learns from different sources. Particularly, in the TSF structure, both the friend and student models are trained using the CPS approach, but the friend model is not supervised by the teacher. %This combination aims to mitigate the issue of over-dependence on the teacher model's performance and provide better diversity for mode training even in later training stage of CPS.

Specifically, the three models are of the same architecture with different weights. The weights in the teacher model are updated using the exponential moving average (EMA) from the student model, while the weights in the student and friend models are updated based on training loss. Based on a teacher-student SSL framework of AEL \cite{hu2021semi}, for labelled data we employ adaptive copy-paste, while unlabelled data use adaptive CutMix. Given both a labelled dataset $\mathcal{A}=\left\{\left(x_i, y_i\right)\right\}_{i=1}^M$ containing $M$ images and an unlabelled data set $\mathcal{B}=\left\{z_i\right\}_{i=1}^N$ with $N$ images, the proposed SSL approach expects to obtain a trained model $Q$ leveraging these labelled and unlabelled data. When working with labelled data, the supervised loss $\mathcal{L}_{sup,s}$ between the $s^\text{th}$ ground truth $y_s$ and its corresponding prediction $p_s$ is defined by using the standard cross-entropy loss function $\ell_{ce}$:
\begin{equation}
\label{equ:equ1}
\mathcal{L}_{sup,s} = \frac{1}{W \times H} \sum_{i=1}^{W \times H}\ell_{c e}\left(p_{i,s}, y_{i,s}\right),
\end{equation}
where $W$ and $H$ refer to the width and height of input images. Both the student and friend models are supervised using the same type of supervised loss function, $\mathcal{L}_{sup,s}$.

Following \cite{sohn2020fixmatch,hu2021semi}, the teacher network generates pseudo-labels $\hat{y}_u^t$ on the $u^\text{th}$ unlabelled image to supervise the student model. The pseudo-labels are improved through an active learning module, called PLAR. The student unsupervised loss $\mathcal{L}_{unsup,u}^{t}$ from teacher-student framework is defined as:
\begin{equation}
\label{equ:equ2}
\mathcal{L}_{unsup,u}^t = \frac{1}{W \times H} \sum_{i=1}^{W \times H}\ell_{c e}\left(r_{i,u}^s, \hat{y}_{i,u}^{t}\right),
\end{equation}
where $r_{i,u}^s$ is the predicted probability map of the $u^\text{th}$ unlabelled image output from the student model.

Similarly, for the $u^\text{th}$ unlabelled image, following CPS \cite{luo2022semi,filipiak2021n}, the student model and its friend model generate pseudo-labels $\hat{y}_u^s$ and $\hat{y}_u^f$ respectively based on their prediction $r_{u}^s$ and $r_{u}^f$ after the $\arg\max()$ function.  These pseudo-labels are used to perform cross-supervision for the two networks.  Thus, the CPS unsupervised loss $\mathcal{L}_{unsup,u}^f$ among student and friend models is defined by using the cross-entropy loss function:
\begin{equation}
\begin{aligned}
\label{equ:equ3}
\mathcal{L}_{unsup,u}^f = \frac{1}{W \times H} \sum_{i=1}^{W \times H}\bigg(\ell_{c e}\left(r_{i,u}^s, \hat{y}_{i,u}^{f}\right)+\ell_{c e}\left(r_{i,u}^f, \hat{y}_{i,u}^{s}\right)\bigg),
\end{aligned}
\end{equation}

The final training objective is written as 
\begin{equation}
\label{equ:equ4}
\mathcal{L}_{s,u} = \mathcal{L}_{sup,s} +   \mathcal{L}_{unsup,u}^t +   \mathcal{L}_{unsup,u}^f.
\end{equation}
%where $\lambda_1$ and $\lambda_2$ are the trade-off weights between these losses.

\subsection{PLAR-based Active Learning}
Unlike general active learning, which selects informative unlabelled images for manual labelling during training loops, the proposed active learning strategy assists semi-supervised learning by correcting erroneous areas in pseudo-labels, as illustrated in Figure~\ref{fig:SSL_framework}. The quality of pseudo-labels in semi-supervised learning is always crucial, but often cannot be guaranteed. In this study, pseudo-labels are generated by applying the $\arg\max()$ function to the predicted probabilities output from the teacher model. PLAR includes an error mask decoder that identifies erroneous areas in pseudo-labels, detailed in Section~\ref{subsec:Error_Mask_Decoder}. Since using manual labels to correct all error areas in pseudo-labels is not cost-effective, we propose an automatic labelling module, PLAR, detailed in Section~\ref{sub:pseudo_label
auto-refinement}. 
 
The proposed approach aims to correct parts of potentially erroneous areas in pseudo-labels based on the similarity between the potentially erroneous regions and the labelled areas in feature space. This process does not require manual labelling, thus preserving the labelling budget while improving the quality of pseudo-labels to enhance the training process. Manual labelling is reserved exclusively for the most challenging and uncertain unlabelled data. The hybrid labelling strategy efficiently reduces the budget of manual labelling in active learning. Rather than relying solely on a subset of unlabelled data selected by traditional active learning methods for manual labelling, this approach utilises all unlabelled data with improved pseudo-labels.

% \begin{figure*}[t]
%     \centering
%     \includegraphics[width=\linewidth]{ active_learning.png}
%     \caption{The proposed active learning structure based on the pseudo-label auto-refinement (PLAR) module for unlabelled data}
%     \label{fig:AL_framework}
% \end{figure*}
 
\subsubsection{Error Mask Decoder (EMD)}
\label{subsec:Error_Mask_Decoder}

Since the widely used probability-based error map, derived from applying a threshold to model predictions (probability prediction for each class), mainly targets object edges, it may not adequately cover the real erroneous areas. Rather than directly applying probability-based model predictions with a threshold to generate the error map, we introduce the EMD module within the proposed PLAR active learning framework. The EMD module is designed to predict the error map of pseudo-labels produced by the teacher model, leveraging the representations of the unlabeled training data. It is a 3-layer convolutional neural network (CNN) with normalisation and activation layers. The EMD module is expected to produce more reasonable and distinct error maps compared to the probability-based error maps for two main reasons: first, it is explicitly trained for error map detection by learning from the differences between the model predictions and the ground truth of the labelled data (an easy binary task); and second, it leverages not only probability-based model predictions but also multi-scale features extracted from the input data. In order to obtain a rich representation of each training sample for EMD input, we concatenate feature maps at multiple levels together with the predicted probability map of the teacher model, resulting in each image (and its pseudo-label) being represented by a 531-channel feature map. Since we adopt DeepLabv3+ as the segmentation network for benchmarking purposes, the feature maps used in EMD input are extracted from the early encoder (first block of ResNet-50) and the late encoder (ASPP block). As a result, the EMD output is the corresponding error probability map $x$ for the pseudo-label generated by the current teacher model. The EMD only uses labelled data for supervision, because the real error maps $y$ can be generated by comparing the ground truth labels (segmentation labels) with the pseudo-labels (created by the teacher model) of the image. During training, the loss $\mathcal{L}_{EMD}$ for EMD supervision is defined by binary cross entropy:

\begingroup
\small
\begin{equation}
\mathcal{L}_{EMD} = \frac{1}{W{\times}H} \sum_{i=1}^{W{\times}H}(y_i \log x_i+\left(1-y_i\right) \log \left(1-x_i\right)).
\end{equation}
\endgroup
EMD operates in an inference mode when generating error maps for unlabelled data.. In order to cover as many potentially erroneous areas as possible, the final error map also incorporates the commonly used probability-based error map, applying a threshold for the probability prediction from the teacher model. Specifically, the final error map combines the error map generated by EMD and the confidence-based error map using the \textit{`OR'} operation. An ablation study for the threshold selection is shown in Section \ref{Sec:ablation}.

Based on the combined error mask, areas outside the mask (i.e. reliable areas) are used in a generally supervised way to supervise the student network according to Equation~\ref{equ:equ2}. The pseudo-labels within erroneous areas will be corrected and improved through a hybrid annotation. Based on the representation of the unlabelled image, some pseudo-labels in the error map are automatically corrected by using the PLAR, which does not consume the annotation budget. The remaining labels require manual labelling within an active learning loop.

\subsubsection{Pseudo-label
Auto-refinement}
\label{sub:pseudo_label
auto-refinement}

\begin{figure}[ht!]
    \centering
    \includegraphics[width=\linewidth]{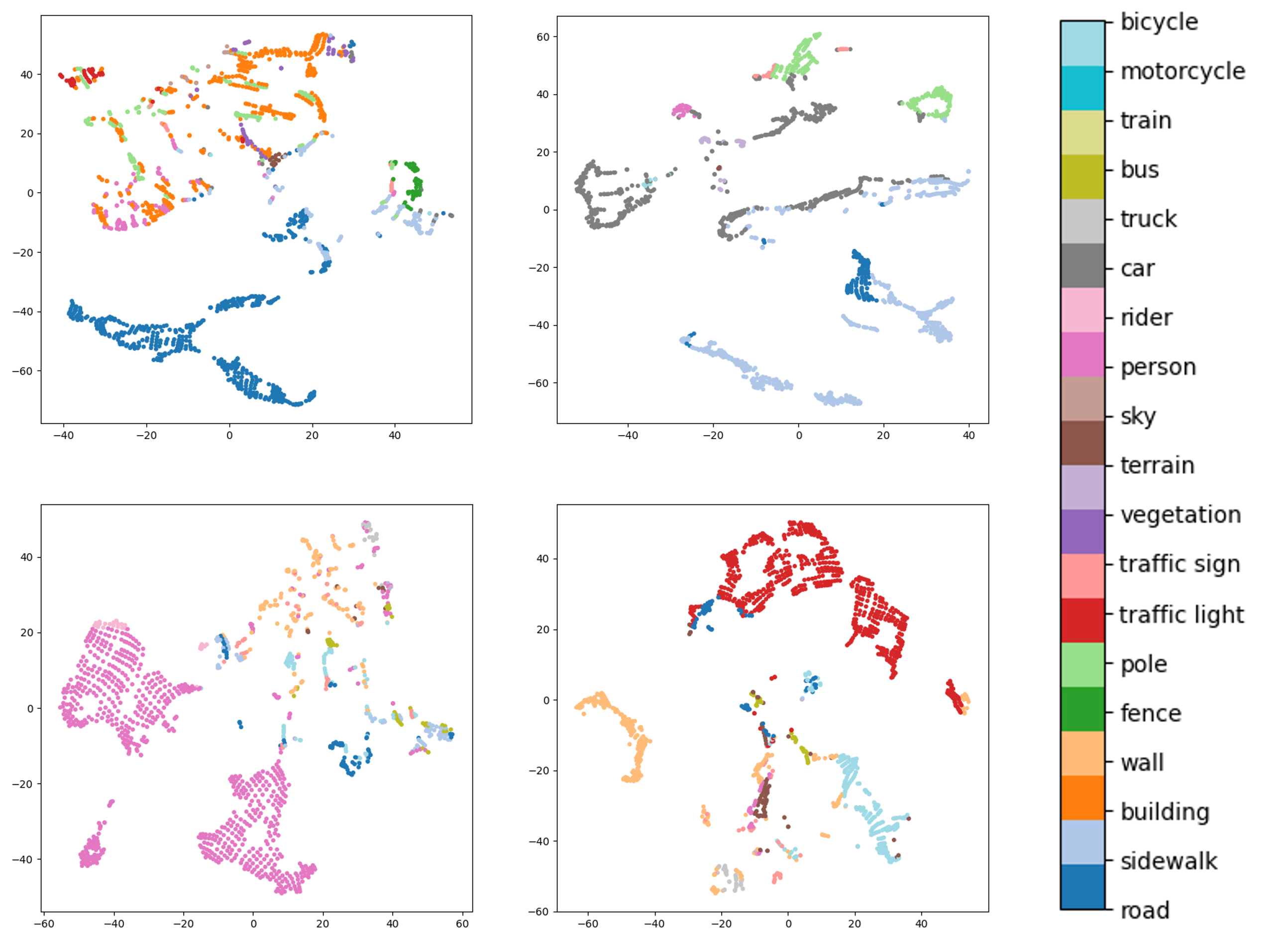}
    \caption{Feature representation of pixels from 4 input images in the potentially erroneous regions of their pseudo-labels using t-SNE. These feature representations %examples of pixels in potential error areas
    demonstrate strong correlations, with most pixels with the same class label (indicated by the same colour in the figures) being allocated into clusters of pixels in T-SNE with similar class labels.}
    \label{fig:Feature_Map}
\end{figure}

In the active learning loop, rather than spending resources on manual labelling to correct pseudo-labels, the proposed pseudo-label auto-refinement attempts to determine whether the existing labels share similar feature characteristics with the new data (which needs to be labelled). This is based on the idea that pixels classified into the same class should exhibit similar representations in the feature space (cluster assumption). Thus, it is feasible to use these existing labels within the same image to correct the erroneous areas of pseudo-labels. The feature maps are extracted from the teacher model at two levels: one from the output of the Atrous Spatial Pyramid Pooling (ASPP) module and the other from the second-to-last layer of the decoder. We define them as $F_1$ and $F_2$.  Figure \ref{fig:Feature_Map} shows visualisations of feature maps corresponding to the pixels in the erroneous regions of pseudo-labels using T-SNE \cite{van2008visualizing}. 

Specifically, the pseudo-label is downsampled to match the length and width of its feature map. Thus, each pixel in a pseudo-label corresponds to a vector in the feature map. We assess the similarity between the features of pixels in the labelled areas and those in the erroneous areas of pseudo-labels using Euclidean and Mahalanobis distances (denoted $E$ and $M$, respectively). For feature map $F_1$, the feature vector of a labelled pixel is defined as $f_1^l \in F_1$ and the feature vector corresponding to a pixel in the erroneous area of the pseudo-label (unlabelled) is denoted as $f_1^e \in F_1$. The Euclidean distance $E_{1c}$ between $f_1^e$ and the mean vector $m$ of the set $F_{1c}^{l}$ including all labelled feature vectors $f_{1c}^l$ corresponding to class $c$ is 
\begin{equation}
E_{1c} = E(f_1^e, m)=\sqrt{\sum_{i=1}^n\left(f_{1i}^e-m_i\right)^2}
\end{equation}
where $n$ is the length of the vector, equal to the channel number of the feature map.

\begin{figure}
    \centering
    \includegraphics[width=0.6\linewidth]{ 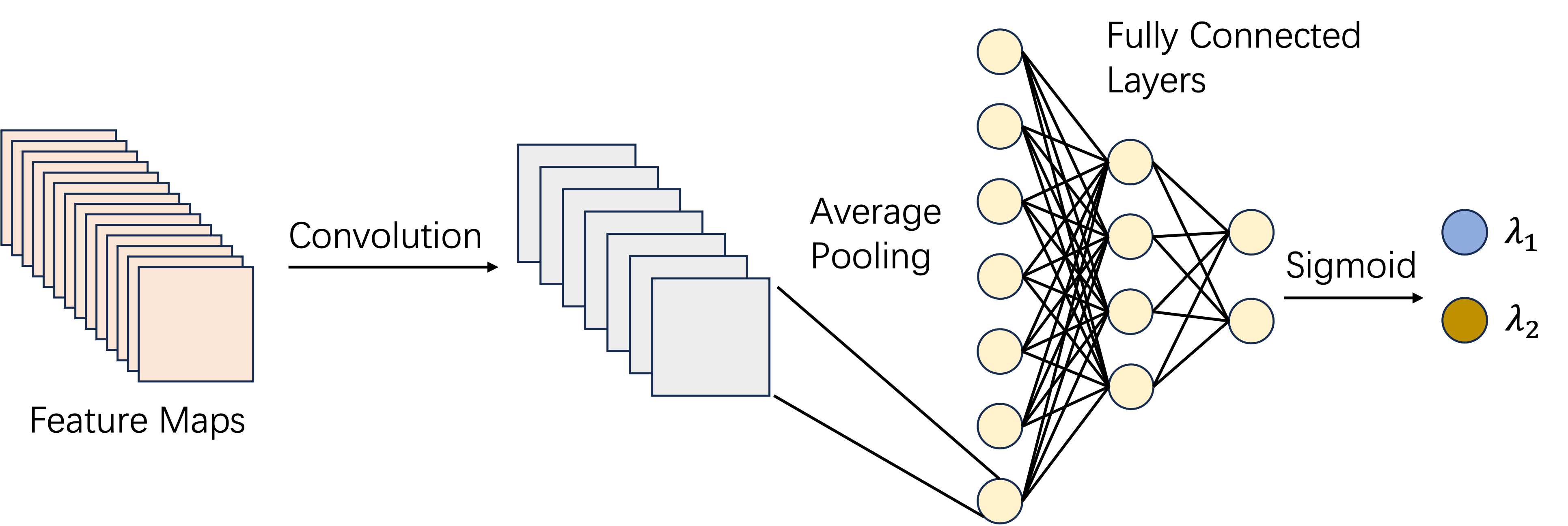}
    \caption{Weight Module. The trade-off weights for computing the hybrid Euclidean and Mahalanobis distance are learnt from the features of the input images.}
    \label{fig:weight_module}
\end{figure}

Similarly, for feature map $F_2$, we got Euclidean distance $E_{2c}$. Thus, the hybrid Euclidean distance $E_c$ corresponding to class $c$ is denoted as
\begin{equation}
    E_c = \lambda_1 E_{1c} + (1-\lambda_1) E_{2c},
\end{equation}
where $\lambda_1$ is a trade-off weight learnt from features by using a weight module shown in Figure \ref{fig:weight_module}. We compute the hybrid Euclidean distance $E_c$ for each class and identify the class corresponding to the smallest Euclidean distance value (across all classes), $E_c^{min}$. The class $c_E$ corresponding to $E_c^{min}$ is considered the potential label for the unlabelled pixel in the erroneous area.

The Mahalanobis distance $M_{1c}$ between the vector $f_1^e$ with the set $F_{1c}^{l}$ containing all labelled feature vectors $f_{1c}^l$ corresponding to class $c$ is 
\begin{equation}
M_{1c}=\sqrt{(f_1^e-m)^T \cdot C^{-1} \cdot(f_1^e-m)},
\end{equation}
where $m$ is the vector of mean values of set $F_{1c}^{l}$ and $C^{-1}$ is inverse covariance matrix of set $F_{1c}^{l}$.

Similarly, for feature map $F_2$, we got Mahalanobis distance $M_{2c}$. Thus, the hybrid Mahalanobis distance $M_c$ corresponding to class $c$ is denoted as
\begin{equation}
    M_c = \lambda_2 M_{1c} + (1-\lambda_2) M_{2c},
\end{equation}
where $\lambda_2$ is a trade-off weight learnt from features by using the weight module in Figure \ref{fig:weight_module}. We compute the hybrid Mahalanobis distance $M_c$ for each class and identify the smallest value, $M_c^{min}$. The class $c_M$ corresponding to $M_c^{min}$ is considered the potential label for the unlabelled pixel in the erroneous area.

So far, each pixel in the erroneous area of the pseudo-label has two potential corrected labels, namely $c_E$ and $c_M$. To ensure label accuracy, we implemented a voting mechanism: if both the Mahalanobis and Euclidean distance metrics indicate the same result, we consider the result to be correct. Specifically, if $c_E$ matches $c_M$, this class is used to correct the erroneous pseudo-labels; otherwise, manual labelling is applied for pseudo-label correction. If the number of pixels requiring manual labelling exceeds the allocated budget, the method will no longer use manual labels and will instead rely on the pseudo-labels. 

\section{Experiments and Results}
\subsection{Datasets}
Given that most state-of-the-art semi-supervised learning (SSL) architectures and active learning (AL) strategies have been initially developed and validated on natural imagery datasets, we begin by benchmarking our proposed method on the widely used CityScapes \cite{cordts2016cityscapes} dataset. This allows for a direct and fair comparison with state-of-the-art techniques in the broader semantic segmentation literature.

Following this initial validation, we extend our evaluation to two established remote sensing benchmarks: ISPRS Vaihingen \cite{rottensteiner2012isprs} and RoadNet \cite{liu2018roadnet}, to assess the adaptability and robustness of our method in remote sensing-specific scenarios. This progression from natural to remote sensing imagery is intentional: it enables us to ground our approach in existing methodological baselines while ultimately demonstrating its effectiveness in the target domain.

% To assess the effectiveness of our methodology across applications, we tested it on three benchmark semantic segmentation datasets: (1) CityScapes \cite{cordts2016cityscapes}, (2) ISPRS Vaihingen \cite{rottensteiner2012isprs} and (3) RoadNet \cite{liu2018roadnet}.

The CityScapes dataset, focused on urban street scenes, includes 2,675 training images, 300 for validation, and 500 for testing, with 19 classes after downsampling to $688\times688$ as in \cite{rangnekar2023semantic,xie2020deal,hu2021semi,wang2022semi_u2pl}. Our TSF framework is evaluated using the same semi-supervised setup as prior methods \cite{hu2021semi, wang2022semi_u2pl, xu2022semi}. The ISPRS Vaihingen dataset, used for land cover classification, includes three-band (NIR, red, green) images at 9cm resolution, DSM data, and six labelled classes. The dataset has 33 patches (average size $2494\times2064$), cropped to $512\times512$ without overlap, resulting in 233, 111, and 398 images for training, validation, and testing, respectively. RoadNet is a benchmark dataset for road network detection with 0.21-m spatial resolution.
It includes RGB images and related road surface maps for the segmentation task. The number of samples for training, validation and testing is 410, 45 and 387, respectively.

\subsection{Implementation Details}
We implemented our methods using the PyTorch framework. To ensure a fair comparison, following \cite{hu2021semi, wang2022semi_u2pl}, the TSF framework employs the same semantic segmentation network as other semi-supervised learning approaches, specifically DeepLabv3+. Also, it utilizes ResNet-101 as the backbone pretrained on ImageNet, while MobileNetv2 serves as the backbone for the active learning approach following \cite{rangnekar2023semantic, xie2020deal}. We train all our networks with a batch size of two and apply the “poly” learning rate strategy and the initial learning rate is set to 0.001 and multiplied by $\left(1-\frac{\text { iter }}{\text { max-iter }}\right)^{0.9}$ at each iteration. For fair comparison with existing methods, the same semantic segmentation architecture (DeepLabv3+) is adopted across all approaches. Consequently, during inference only a single segmentation network is used, and the additional teacher and friend networks are not required. Therefore, the inference-time computational cost, including Parameters, FLOPs, and GPU memory consumption, remains identical to the baseline segmentation model and comparable methods. Specifically, the computational complexity of the model is shown in Table \ref{tab:model_specs}.

\begin{table}[h!]
\centering

\begin{tabular}{l r l}
\hline
\textbf{Specification} & \textbf{Value} & \textbf{Unit} \\
\hline
Input Resolution & 512 $\times$ 512 & pixels \\
Parameters & 40.35 & M \\
FLOPs & 69.45 & GFLOPs \\
Forward Pass RAM & 260.00 & MB \\
\hline
\end{tabular}
\caption{Detailed specifications of the semantic segmentation model (DeepLabv3+) in the inference phase.}
\label{tab:model_specs}
\end{table}

\subsection{SSL Structure Comparison}

We evaluated the proposed TSF semi-supervised learning framework on the CityScapes dataset with a comparative study to the recent approaches of MT\cite{tarvainen2017mean}, CutMix\cite{yun2019cutmix}, CCT\cite{ouali2020semi}, GCT\cite{ke2020guided}, CPS\cite{chen2021semi}, AEL\cite{hu2021semi}, PS-MT\cite{liu2022perturbed}, U2PL\cite{wang2022semi_u2pl}, and UniMatch\cite{yang2023revisiting}. We split the entire training dataset into 1/16, 1/8, 1/4, and 1/2 ratios, following the approach in \cite{hu2021semi, wang2022semi_u2pl}. 

\begingroup
\setlength{\tabcolsep}{1pt}
\begin{table}[htbp]
  \renewcommand{\arraystretch}{1}\centering
  \caption{Performance comparison of state-of-the-art SSL approaches on CityScapes validation set by mIoU (\%). PS-MT was not evaluated at the ratio of 1/16, and no checkpoint was shared for that, so the result of PS-MT for the 1/16 ratio is unavailable. Built upon AEL, the proposed TSF shows its improvement over AEL in the final row of the table.}
    \begin{tabular}{p{4.5cm}p{2.5cm}p{2.5cm}p{2.5cm}p{2.5cm}}
\cmidrule{1-5}          & \multicolumn{1}{l}{1/16 (186) } & \multicolumn{1}{l}{1/8 (372) } & \multicolumn{1}{l}{1/4 (744) } & \multicolumn{1}{l}{1/2 (1488)}   \\
\cmidrule{1-5}    SupOnly  & 65.74 & 72.53 & 74.43 & 77.83   \\
    MT \cite{tarvainen2017mean}  & 69.03 & 72.06 & 74.2  & 78.15   \\
    CutMix \cite{yun2019cutmix}& 67.06 & 71.83 & 76.36 & 78.25   \\
    CCT  \cite{ouali2020semi} & 69.32 & 74.12 & 75.99 & 78.1    \\
    GCT \cite{ke2020guided}  & 66.75 & 72.66 & 76.11 & 78.34   \\
    CPS \cite{chen2021semi} & 69.78 & 74.31 & 74.58 & 76.81   \\
    AEL \cite{hu2021semi} & 74.45 & 75.55 & 77.48 & 79.01  \\
    % U2PL (w/ CutMix)  & 70.30  & 74.37 & 76.47 & 79.05 &  \\
    PS-MT \cite{liu2022perturbed} & -- &76.89& 77.60& 79.09\\
    U2PL \cite{wang2022semi_u2pl} & 74.90  & 76.48 & 78.51 & 79.12  \\
    UniMatch \cite{yang2023revisiting} &\textcolor[rgb]{ 1,  0,  0}{76.60}&	\textcolor[rgb]{ 1,  0,  0}{77.90}&	79.20&	79.50\\
    AEL-based TSF & 75.59 & 77.86 & \textcolor[rgb]{ 1,  0,  0}{79.5} & \textcolor[rgb]{ 1,  0,  0}{80.32}   \\\hline
    \textcolor{olive}{Gain ($\triangle$) to AEL} & \textcolor{olive}{$\uparrow$ 1.14}& \textcolor{olive}{$\uparrow$ 2.31 }& \textcolor{olive}{$\uparrow$ 2.02 }& \textcolor{olive}{$\uparrow$ 1.31}\\
\cmidrule{1-5}    \end{tabular}%
  \label{res:semi-supervised_learning}
\end{table}%
\endgroup

Table \ref{res:semi-supervised_learning} presents a performance comparison of the proposed TSF SSL method with state-of-the-art semi-supervised learning approaches at each ratio, using the widely adopted segmentation metric, mean Intersection over Union (mIoU). It is important to highlight that the results in Table 1 pertain solely to SSL frameworks, which do not include the active learning component. In the case, when no unlabelled data is utilised, the performance of the supervised baseline model, SupOnly, is not satisfactory across all data partition protocols shown by Table \ref{res:semi-supervised_learning}. The state-of-the-art semi-supervised learning approaches MT, CutMix, CCT, GCT, and CPS, show some improvement compared to SupOnly, though it is not substantial. On the other hand, the AEL semi-supervised learning approach achieved a significant improvement using a teacher-student framework, greatly enhancing segmentation performance even with only 1/16th of the labelled data. One of the most recent semi-supervised learning approaches used in this study, UniMatch, demonstrated comparable performance to our proposed STF framework, with UniMatch slightly outperforming STF at the 1/16 and 1/8 ratios. However, STF outperformed UniMatch along with all other state-of-the-art methods reported in the table at the other two ratios.

It is important to note that, considering the proposed method builds upon AEL\cite{hu2021semi}, Table \ref{res:semi-supervised_learning} also highlights the performance improvements relative to AEL in the final row. Our method provides clear and consistent improvements to AEL where there is more than a 2\% mIoU improvement for the 1/8 and 1/4, and over 1\% for the remaining two ratios. Compared to the SupOnly baseline, our method achieves a 9.85\% improvement at the 1/16 ratio, where the training data is very limited. Our method also significantly outperforms the existing state-of-the-art methods in 1/4 and 1/2 ratios.

\begin{table}[htbp]
\centering
\caption{Comparison of memory usage and execution time between the teacher-student baseline (AEL) and the proposed TSF network.}
\label{tab:performance_comparison}
\begin{tabular}{ccc}
\toprule
\textbf{Method} & \textbf{Memory Usage (GB)} & \textbf{Time (seconds)} \\
\midrule
AEL   & 4.14 & 31.12 \\
Ours  & 7.69 & 45.66 \\
\bottomrule
\end{tabular}
\end{table}

Since the TSF framework involves an additional model (Friend) during the training phase, this design introduces extra training overhead. Table \ref{tab:performance_comparison} presents memory usage and training time (100 iterations) between the teacher-student baseline (AEL) and the proposed TSF network. While our method exhibits a higher memory footprint (7.69 GB vs. 4.14 GB) and increased training time (45.66 s vs. 31.12 s), these costs remain within the operational limits of common hardware. This additional computational cost results in notable performance gains, allowing the method to reach state-of-the-art performance. 

The results show that the proposed TSF framework is efficient, consistent and highly competitive, making it a practical choice for incorporating active learning to further reduce labelling costs while enhancing performance.

\subsection{Results for the Proposed Hybrid Model}
%The proposed active learning method was evaluated on two datasets: the CityScape dataset for natural image semantic segmentation and the Vaihingen dataset for land cover classification in remote sensing images. 
To the best of our knowledge, this is the first study to evaluate active learning approaches for semantic segmentation across both natural images and remote sensing imagery. The proposed methodology (referred to as `Ours' in tables and figures), which combines the TSF framework with active learning strategies, was previously illustrated in Figure \ref{fig:SSL_framework}. We assessed its performance against leading active learning methods, including Core-Set \cite{sener2017active}, VAAL \cite{sinha2019variational}, QBC \cite{seung1992query}, and DEAL \cite{xie2020deal}, across multiple datasets. The S4AL  \cite{rangnekar2023semantic} and SegXAL \cite{mandalika2024segxal} were evaluated only on the CityScapes dataset, as the authors have not shared the code, making them non-reproducible for remote sensing datasets in this study. For the proposed method, we limit the labelled data to 16\% for the CityScapes dataset to match the ratio used by S4AL (other methods use 40\% labelled data), ensuring a fair comparison by aligning with the method that uses the smallest amount of labelled data. We did not impose a limit on the use of labelled data, and the training process gradually consumes some labelled data for the most uncertain area in each iteration; by the end of the active learning training iterations, 33\% of the labelled data was automatically utilised for both the Vaihingen dataset and RoadNet datasets.

\begin{table}
    \centering\setlength{\tabcolsep}{1.5pt}
        \caption{Performance comparison with the state-of-the-art methods on the CityScapes validation set with a metric of IoU for each class and mIoU (\%). The percentages in the ``Label" column indicate the proportion of labels used in the method relative to the full labels. The best results are highlighted in \textbf{bold}. *We have capped the label budget at 16\% to match the method with the least label use.}
    \label{tab:results_cityscope}
    \begin{tabular}{@{}cccccccccccc@{}}
\toprule Method & Label \% & Road & 
\makecell[l]{Side \\ walk}& Building & Wall & Fence & Pole & \makecell[l]{Traffic \\ Light} & \makecell[l]{Traffic \\ Sign} & Vegetation & Terrain \\
\toprule Random & 40\%& 96.03 & 72.36 & 86.79 & 43.56 & 44.22 & 36.99 & 35.28 & 53.87 & 86.91 & 54.58 \\%\hline 
Entropy & 40\%& 96.28 & 73.31 & 87.13 & 43.82 & 43.87 & 38.10 & 37.74 & 55.39 & 87.52 & 53.68 \\%\hline 
Core-Set \cite{sener2017active} & 40\% & 96.12 & 72.76 & 87.03 & 44.86 & 45.86 & 35.84 & 34.81 & 53.07 & 87.18 & 53.49 \\%\hline 
VAAL \cite{sinha2019variational} & 40\% &  96.22 & 73.27 & 86.95 & 47.27 & 43.92 & 37.40 & 36.88 & 54.90 & 87.10 & 54.48\\%\hline 
QBC \cite{seung1992query} & 40\% & 96.07 & 72.27 & 87.05 & 46.89 & 44.89 & 37.21 & 37.57 & 54.53 & 87.51 & 55.13 \\%\hline 
DEAL \cite{xie2020deal} & 40\% & 95.89 & 71.69 & 87.09 & 45.61 & 44.94 & 38.29 & 36.51 & 55.47 & 87.53 & 56.90 \\%\hline 
SegXAL \cite{mandalika2024segxal} & 40\%& 96.98 & 73.43 & 88.34 & 46.88 & 45.38 & 36.12 & 37.36 & 55.38 & 87.84 & 59.87 \\
S4AL \cite{rangnekar2023semantic} &16\%    & 97.73 & \textbf{81.76} & 88.63 & \textbf{51.42} & 47.40 & 36.00 & 43.91 & 58.27 & 89.72 & \textbf{62.01} \\\toprule
Ours &16\%* & \textbf{97.44}   & 79.74   & \textbf{89.12}   & 45.21   & \textbf{50.66}   & \textbf{45.11}   & \textbf{53.04}   & \textbf{66.32}   & \textbf{90.17}   & 59.68   \\\bottomrule \bottomrule
& Label \% &Sky & 
\makecell[l]{Pedes- \\ trian}  & Rider & Car & Truck & Bus & Train & \makecell[l]{Motor \\ Cycle} & Bicycle & $\mathrm{mIoU}$ \\\toprule 
Random &40\% & 91.47 & 62.74 & 37.51 & 88.05 & 56.64 & 61.00 & 43.69 & 30.58 & 55.67 & 59.00 \\%\hline 
Entropy & 40\%& 92.05 & 63.96 & 34.44 & 88.38 & 59.38 & 64.64 & 50.80 & 36.13 & 57.10 & 61.46 \\%\hline 
Core-Set \cite{sener2017active} & 40\% & 91.89 & 62.48 & 36.28 & 87.63 & 57.25 & 67.02 & \textbf{56.59} & 29.34 & 53.56 & 60.69 \\%\hline 
VAAL \cite{sinha2019variational} & 40\%& 91.63 & 63.44 & 38.92 & 87.92 & 50.15 & 63.70 & 52.36 & 35.99 & 54.97 & 60.92 \\%\hline 
QBC \cite{seung1992query} & 40\% & 91.87 & 63.79 & 38.76 & 88.04 & 53.88 & 65.92 &  54.32  & 32.68 & 56.15 & 61.29 \\%\hline 
DEAL \cite{xie2020deal} & 40\%  & 91.78 & 64.25 & 39.77 & 88.11 & 56.87 & 64.46 & 50.39 & 38.92 & 56.59 & 61.64 \\%\hline 
SegXAL \cite{mandalika2024segxal} & 40\%& 92.93 & 62.56 & 39.07 & 88.11 & 59.47 & 65.70 & 46.88 & 35.53 & 54.71 & 65.11 \\
S4AL \cite{rangnekar2023semantic} &16\% & 92.81 & 65.62 & 39.71 & 90.52 & \textbf{66.07} & 65.31 & 46.03 & \textbf{46.88} & 61.77 & 64.80 \\\toprule
Ours &16\%* & 
\textbf{93.01}    & \textbf{70.41}    & \textbf{46.42}    & \textbf{91.78}    & 53.58    & \textbf{70.53}    & 42.99    & 44.97    & \textbf{67.94}    & \textbf{66.22}\\
\bottomrule
\end{tabular}
\end{table}

\begin{table}
    \centering
        \caption{Performance comparison with the state-of-the-art on the Vaihingen dataset by IoU for each class and mIoU (\%). The percentages in the ``Label" column indicate the proportion of labels used in the method relative to the full labels. The best results are highlighted in \textbf{bold}. *By the end of the active learning training iteration, 33\% of the labelled data was used.}
    \label{tab:results_vaihingen}
    \begin{tabular}{@{}ccccccccc@{}}

\toprule Method & Label \% & Road & 
\makecell[c]{Impervious \\ surfaces}& \makecell[c]{Low \\vegetation } & Tree & Car & \makecell[c]{Clutter/ \\ background} &mIoU \\  
\toprule Random & 40\%& 71.38 &	74.04 &	56.95 &	68.98 &	42.30 &	11.65 & 54.22\\
 Entropy & 40\%&72.27 &76.30 &53.53 &67.19 &46.27 &2.51& 53.01  \\
 Core-Set \cite{sener2017active} & 40\%& 73.31 &77.84 &58.00 &69.89 &45.47 &10.33 & 55.81  \\
 VAAL \cite{sinha2019variational} & 40\% &  73.29 &77.90 &55.88 &68.48 &45.93 &13.21 & 55.78 \\
 QBC \cite{seung1992query} & 40\%& 72.44 &76.35 &54.51 &67.96 &47.24 &9.75 & 54.71\\
 DEAL \cite{xie2020deal} & 40\%& 72.59 &	77.04&	55.09&	68.62&	39.72&	18.31 & 55.23\\
\toprule Ours &33\%* & \textbf{77.15}   & \textbf{82.78}   & \textbf{63.07}   & \textbf{74.13}   & \textbf{53.50}   & \textbf{27.89}   & \textbf{63.09} \\
\bottomrule
\end{tabular}
\end{table}

\begin{table}
    \centering
        \caption{Performance comparison with the state-of-the-art on the RoadNet dataset by IoU for each class and mIoU (\%). The percentages in the ``Label" column indicate the proportion of labels used in the method relative to the full labels. The best results are highlighted in \textbf{bold}. *By the end of the active learning training iteration, 33\% of the labelled data was used.}
    \label{tab:results_roadnet}
    \begin{tabular}{p{4cm}p{2.5cm}p{2.5cm}p{2.5cm}p{2.5cm}}
\toprule Method & Label \% & Background & Road &  mIoU  \\  
\toprule Random & 40\% & 94.70 &69.55  &82.13 \\
Entropy & 40\% & 94.79  & 69.77   & 82.28 \\
Core-Set \cite{sener2017active} & 40\%& 94.99  & 69.72  & 82.35 \\
VAAL \cite{sinha2019variational} & 40\%  & 94.73  & 68.92  & 81.82  \\
QBC \cite{seung1992query} & 40\% & 94.74  & 69.79  & 82.26 \\
DEAL \cite{xie2020deal} & 40\%  &  95.78  &  74.31  &  85.04 \\
\toprule Ours &33\%*&\textbf{96.98} & \textbf{80.52} & \textbf{88.75} \\
\bottomrule
\end{tabular}
\end{table}

\subsubsection{Results on CityScapes} 
%We evaluated the performance of the proposed method against leading active learning approaches, including Core-Set \cite{sener2017active}, VAAL \cite{sinha2019variational}, QBC \cite{seung1992query}, DEAL \cite{xie2020deal}, and S4AL \cite{rangnekar2023semantic} on CityScape dataset, 
Table \ref{tab:results_cityscope} presents the performance analysis, showing the Intersection over Union (IoU) for each class and the mean IoU (mIoU) across all classes. The proposed method and S4AL utilise 16\% of the pixel-wise labels and all unlabelled images, while the other methods use 40\% of the labelled training data from CityScapes. The proposed method achieves the highest overall mIoU of 66.22\%, outperforming all other state-of-the-art methods, including those using 40\% of the data. A detailed per-class analysis reveals that the proposed method achieves the highest IoU in several key classes: Building (89.12\%), Fence (50.66\%), Pole (45.11\%), Traffic Light (53.04\%), Traffic Sign (66.32\%), Vegetation (90.17\%), Sky (93.01\%), Side Walk (70.41\%), Rider (46.42\%), Car (91.78\%), Bus (70.53\%), Bicycle (67.90\%). Among the methods that utilise 40\% of the labelled data, the mIoU performance is relatively similar, with DEAL and Entropy standing out with slightly higher mIoUs of 61.64\% and 61.46\%, respectively. The Random method generally yields lower performance across most classes, which aligns with expectations, as random sampling does not produce class-specific IoU improvements. This further illustrates the importance of targeted sampling strategies in enhancing segmentation performance.

\subsubsection{Results on ISPRS Vaihingen} 
Table \ref{tab:results_vaihingen} presents a performance comparison of various leading methods on the land cover classification dataset, Vaihingen, evaluated using Intersection over Union (IoU) for each class and the mean IoU (mIoU). %The results for S4AL are not included in the table because its code was not made available to reproduce results for the remote sensing dataset. 
The proposed method demonstrates greater strength for the remote sensing dataset and achieves the highest mIoU of 63.09\%, outperforming all other methods, despite utilising only 33\% of the labelled training data. For per-class IoU, the proposed method consistently outperforms the other methods across most individual classes. Notably, it achieves the highest IoU in several critical classes: Road (77.15\%), Impervious Surfaces (82.78\%), Low Vegetation (63.07\%), Tree (74.13\%), Car (53.50\%), Clutter/Background (27.89\%). Among the methods that utilise 40\% of the labelled data, ``Core-Set" and ``VAAL" show competitive performances, with mIoU of 55.81\% and 55.78\%, respectively. However, they fall short of the performance achieved by the proposed method. This significant result underscores the effectiveness of our approach in leveraging a smaller proportion of labelled data while achieving superior segmentation accuracy. 

\subsubsection{Results on RoadNet Dataset} 

Table \ref{tab:results_roadnet} presents the performance comparison on the RoadNet dataset in terms of class-wise IoU and mean IoU (mIoU). All baseline methods (Random, Entropy, Core-Set, VAAL, QBC, and DEAL) use 40\% of the labelled training data. In contrast, our proposed PLAR active learning approach uses only 33\% of the training data by the end of training iterations. Despite this significant reduction in labelled data, our method outperforms all comparison methods by a considerable margin. Specifically, our approach achieves 96.98\% IoU for the background class and 80.52\% for the road class, both substantially higher than the best baseline results (95.78\% and 74.31\%, respectively). In terms of mIoU, our method reaches 88.75\%, outperforming the strongest baseline (DEAL) by nearly 4 percentage points, even though the baselines utilise more labelled data.

These results demonstrate that our method effectively identifies informative samples for manual annotation, which significantly reduces the annotation cost while achieving state-of-the-art performance. Moreover, its effectiveness on remote sensing datasets highlights its potential for real-world semantic segmentation applications where labelled data is limited.

\subsection{Qualitative Analysis}
\begin{figure}
    \centering
    \includegraphics[width=0.7\linewidth]{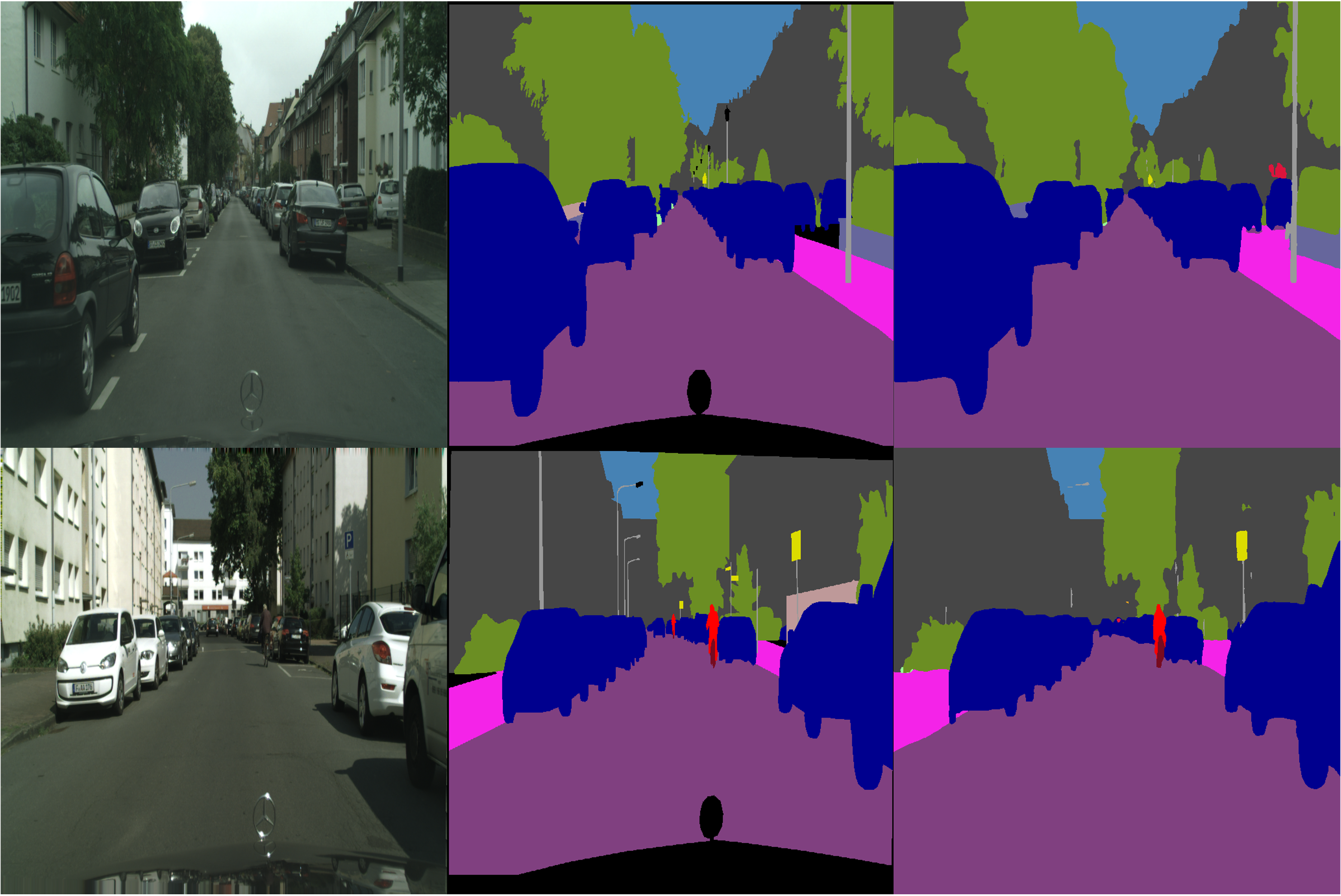}
    \caption{Visualisation results of the proposed approach compared to the ground truth. The images, displayed from left to right, represent RGB images, ground truth, and predictions.}
    \label{fig:vis_cityscopes}
\end{figure}

As shown in Figure \ref{fig:Visual_results}, our proposed method demonstrates superior performance in both land cover classification (Vaihingen dataset) and road network detection (RoadNet dataset). Visually, our results are closer to the ground truth, with fewer misclassifications and better structural preservation. Quantitatively, we consistently achieve the highest mIoU scores across all examples, outperforming state-of-the-art active learning methods such as VAAL, Core-Set, and DEAL. These results highlight the effectiveness and generalizability of our approach for different remote sensing tasks. Since visualisations of many benchmark methods on the Cityscapes dataset are not publicly available, and rerunning their code may lead to inconsistent results, we focus on presenting qualitative comparisons between our method and the ground truth in Figure \ref{fig:vis_cityscopes}.

\begin{figure}
    \centering
    \includegraphics[width=\linewidth]{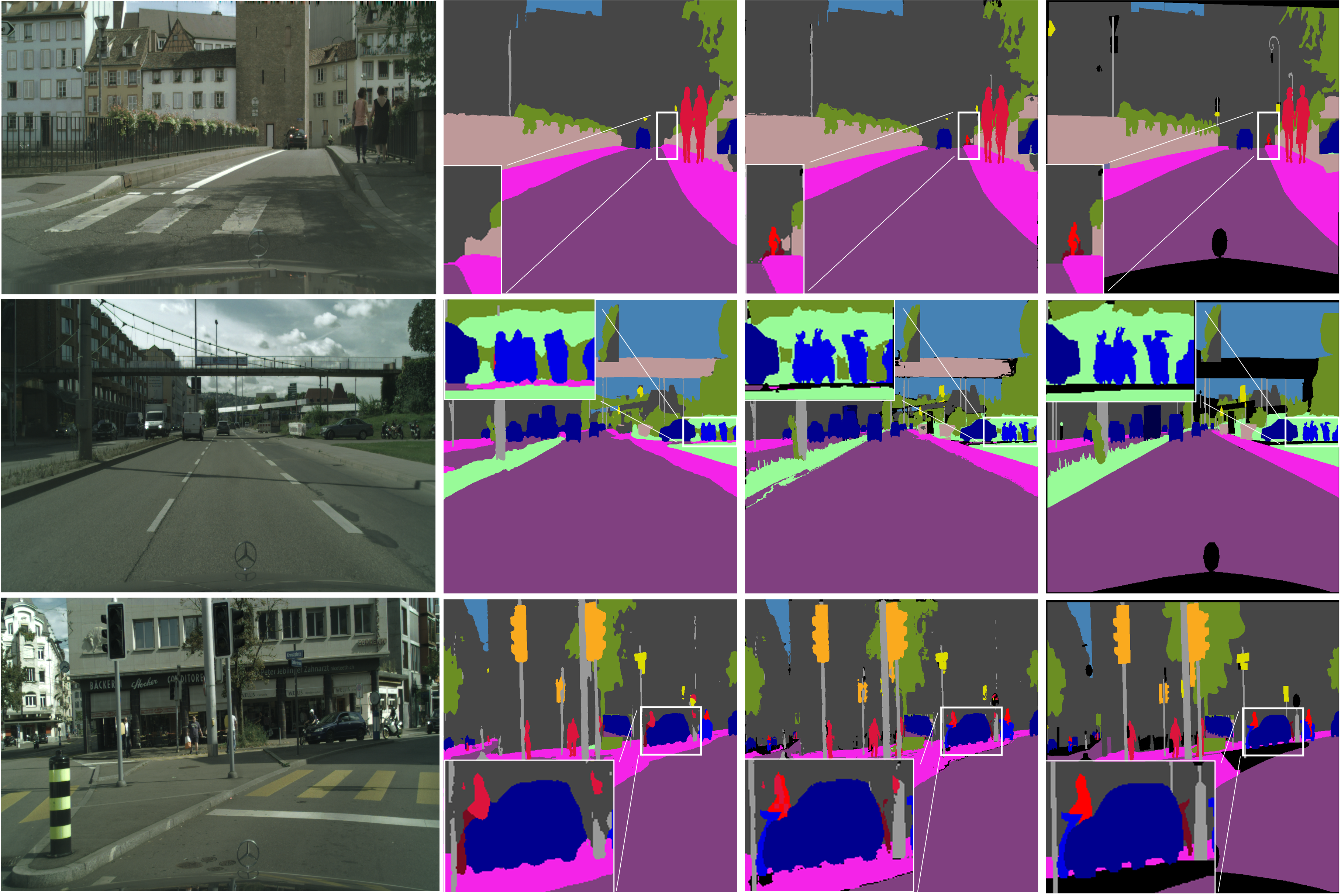}
    \caption{Visualisation of pseudo-label correction. The images, displayed from left to right, represent RGB images, pseudo-labels, corrected pseudo-labels, and ground truth.}
\label{Fig:pseudo_label_correction}
\end{figure}

To utilise the Cityscapes dataset for visualisation, we illustrate the performance of pseudo-label correction in Figure \ref{Fig:pseudo_label_correction}. The improved pseudo-labels exhibit much greater accuracy in handling details than the original version, closely aligning with the ground truth. These improved pseudo-labels, therefore, offer more effective supervision for models learning from labelled data.

\afterpage{
  \clearpage
  \begin{landscape}
\begin{figure}[p]
\centering \setlength \tabcolsep{3pt}
    \begin{tabular}{m{0.1\linewidth}m{0.1\linewidth}m{0.1\linewidth}m{0.1\linewidth}m{0.1\linewidth}m{0.1\linewidth}m{0.1\linewidth}m{0.1\linewidth}m{0.1\linewidth}c}
         \centering Image & \centering GT & \centering Ours & \centering DEAL & \centering QBC   & \centering VAAL & \centering Core-Set & \centering Entropy& \centering Random & \\ 
        \includegraphics[width=\linewidth]{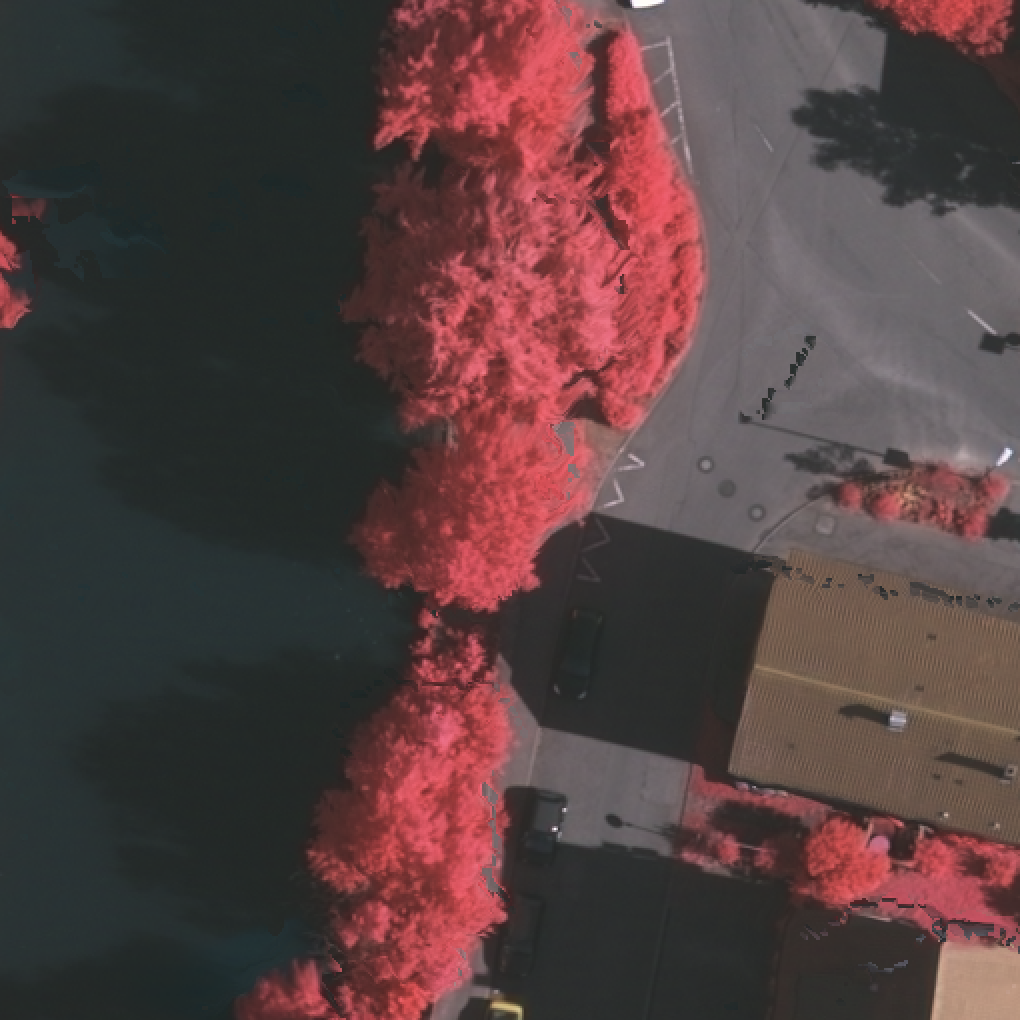} & \includegraphics[width=\linewidth]{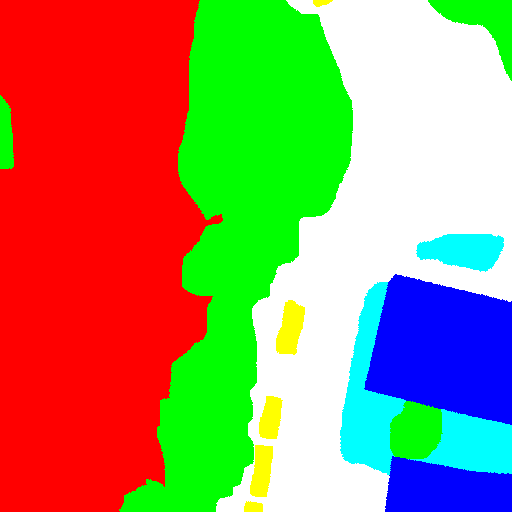} & \includegraphics[width=\linewidth]{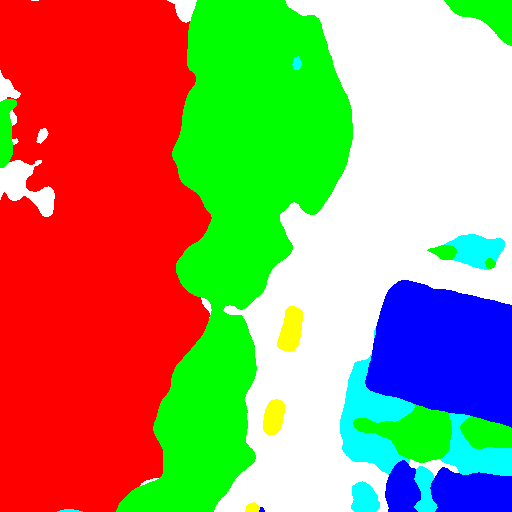} & \includegraphics[width=\linewidth]{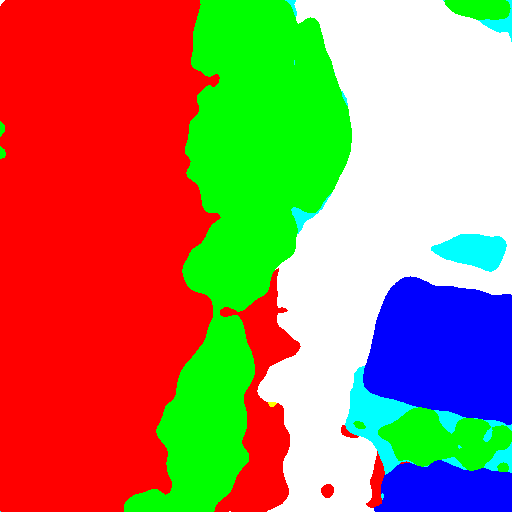} & \includegraphics[width=\linewidth]{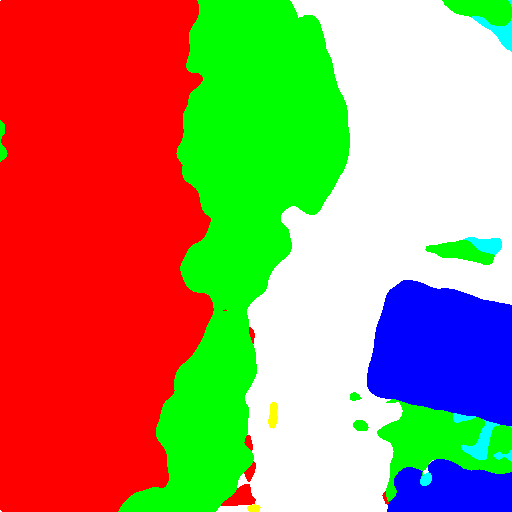} & \includegraphics[width=\linewidth]{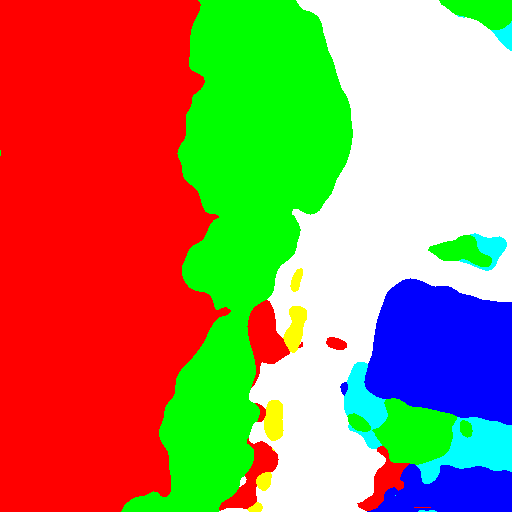} & \includegraphics[width=\linewidth]{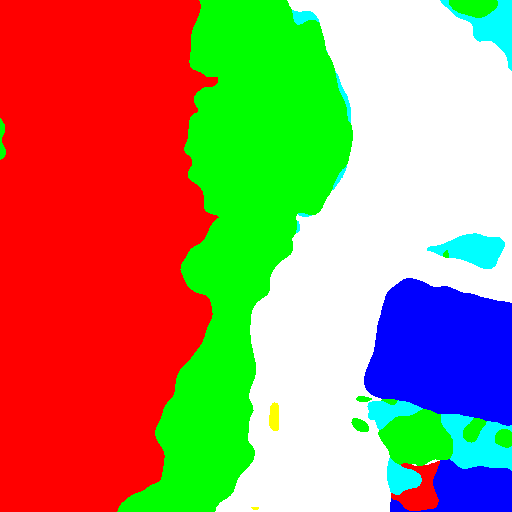} & \includegraphics[width=\linewidth]{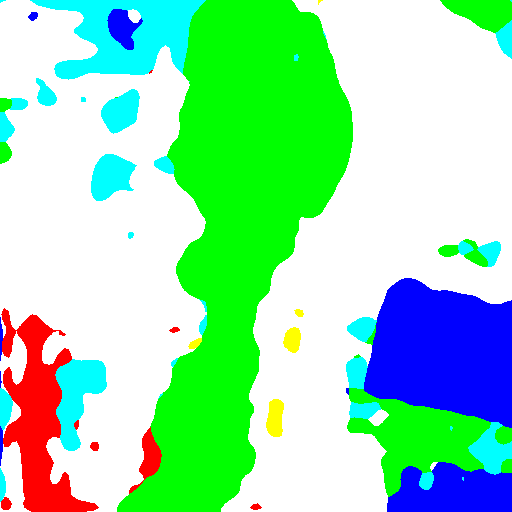} & \includegraphics[width=\linewidth]{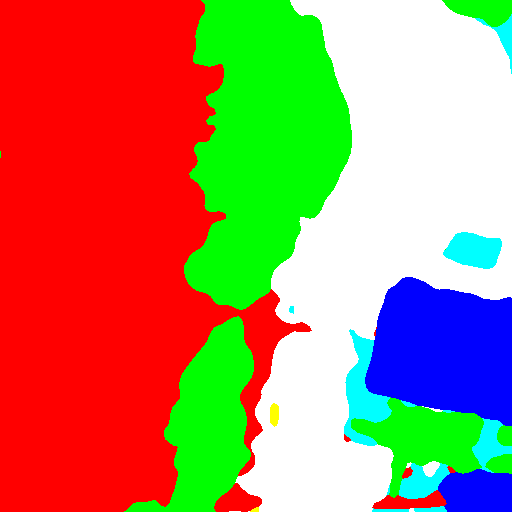}  &\\

&   & \centering	 \textbf{75.65\%} & \centering	 64.41\% & \centering 62.37\% & \centering	73.00\% & \centering	 65.31\% & \centering	 44.29\% & \centering	 62.22\% & \\
         \includegraphics[width=\linewidth]{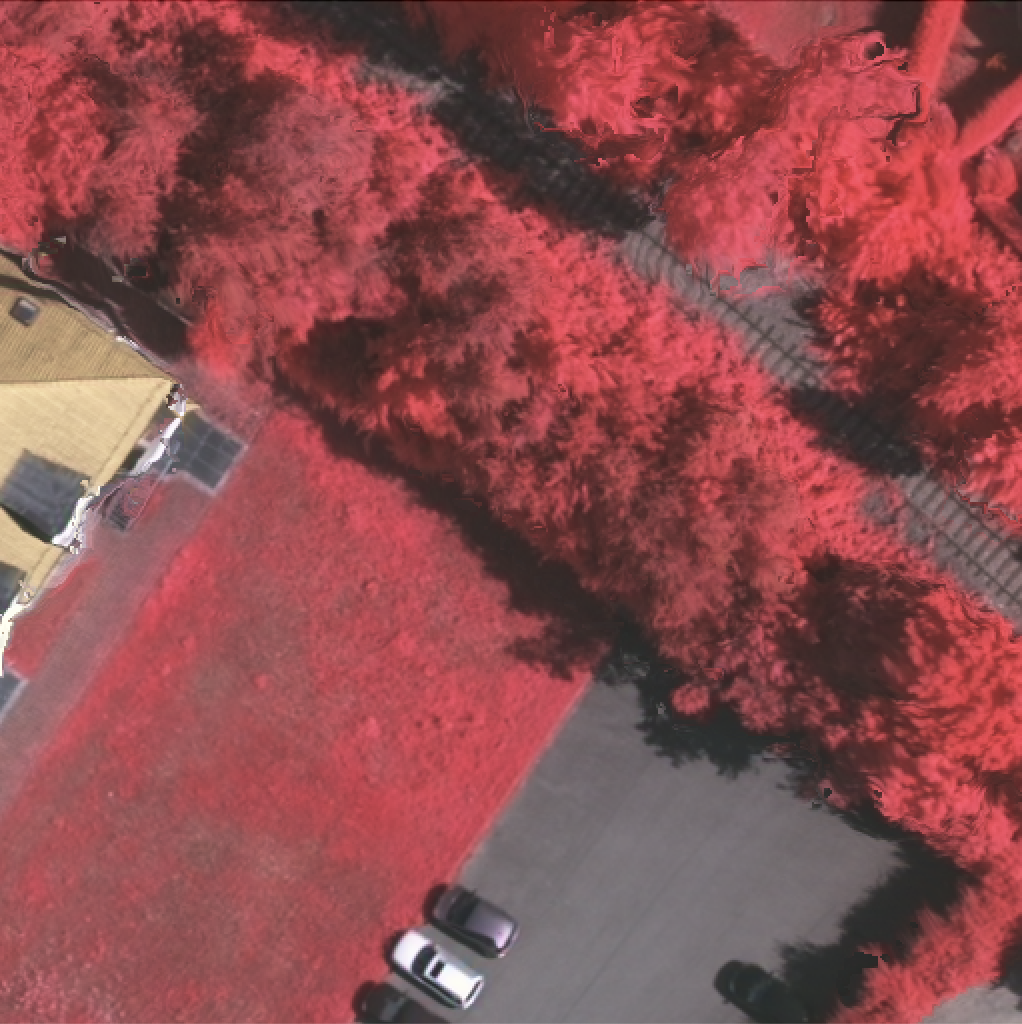} & \includegraphics[width=\linewidth]{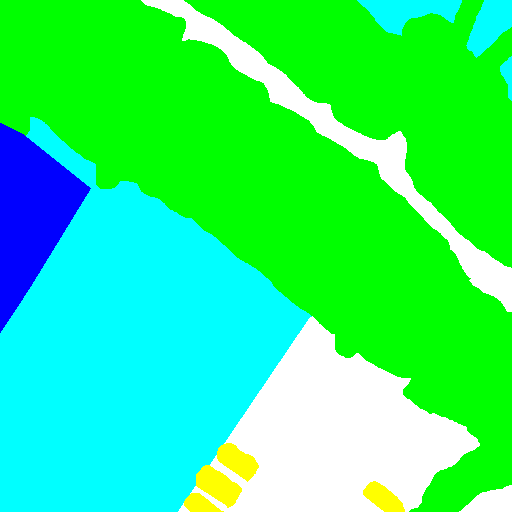} & \includegraphics[width=\linewidth]{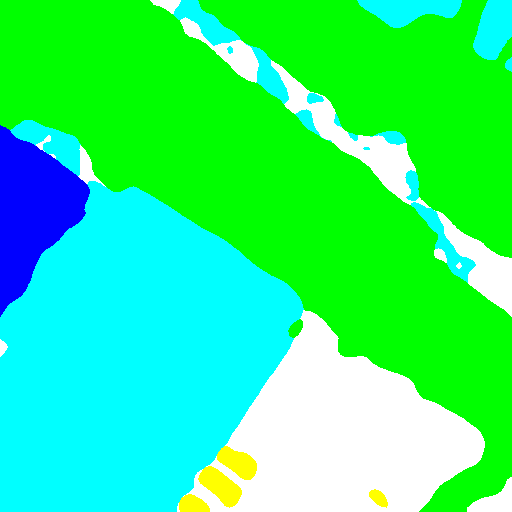} & \includegraphics[width=\linewidth]{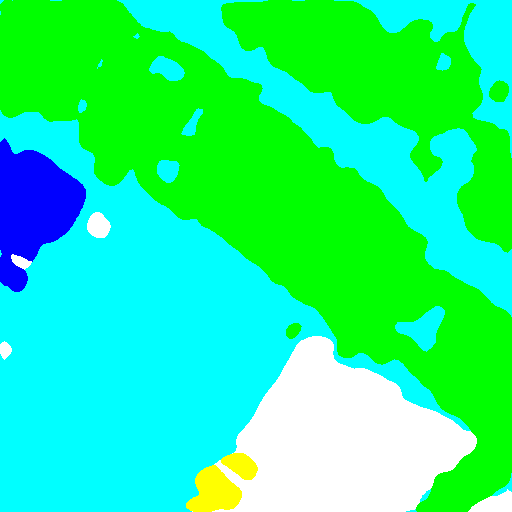} & 
        \includegraphics[width=\linewidth]{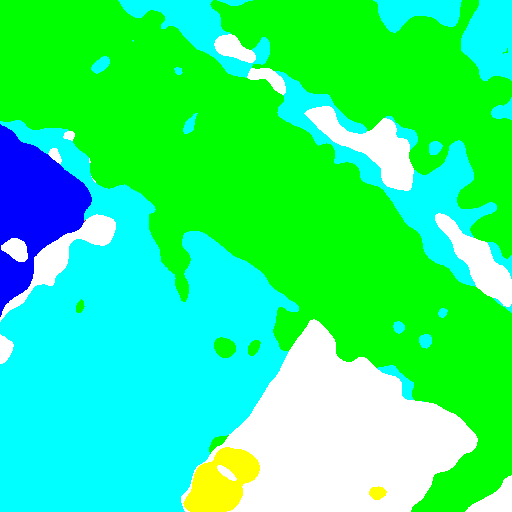} & \includegraphics[width=\linewidth]{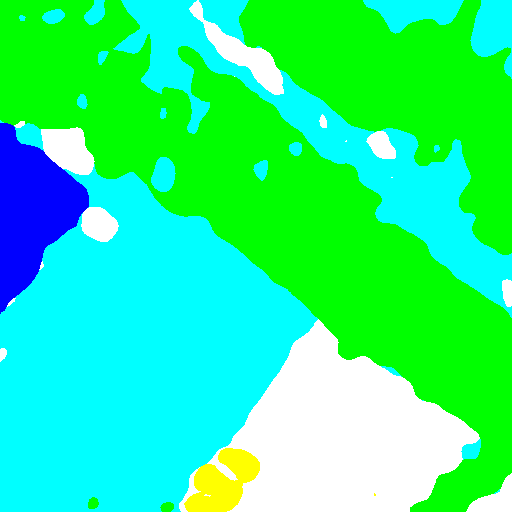} & \includegraphics[width=\linewidth]{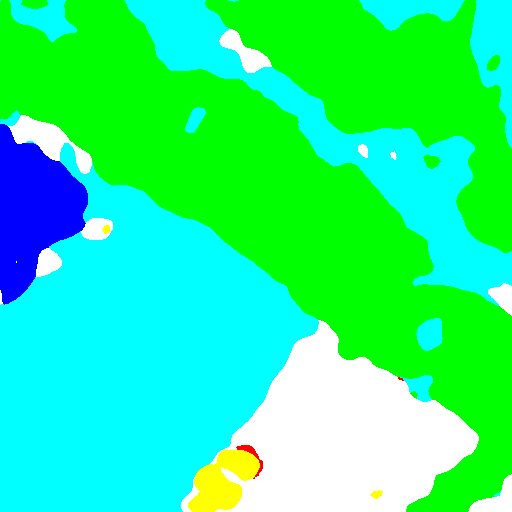} & \includegraphics[width=\linewidth]{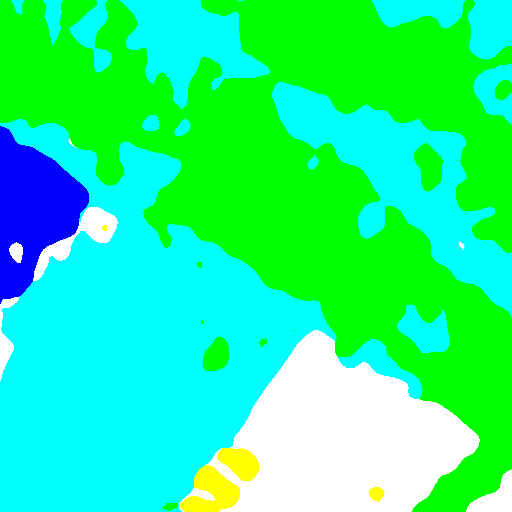} & \includegraphics[width=\linewidth]{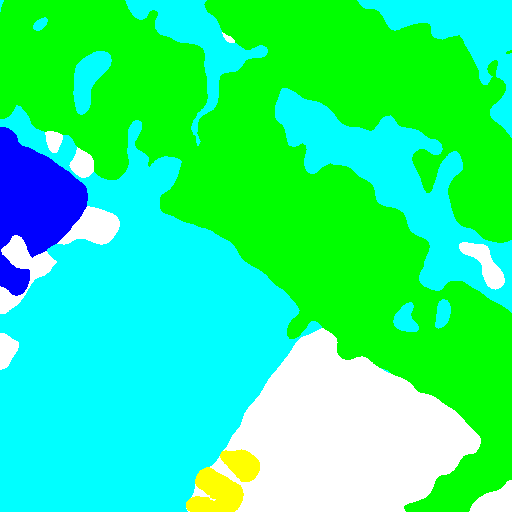}  \\

          &   & \centering	 \textbf{69.65\%} & \centering	 57.77\% & \centering	 63.11\% & \centering	 61.64\% & \centering	 60.81\% & \centering	 58.71\% & \centering	 58.17\% & \\

        \includegraphics[width=\linewidth]{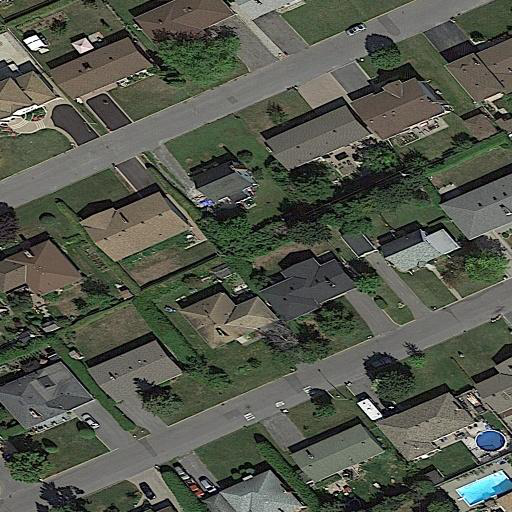} & \includegraphics[width=\linewidth]{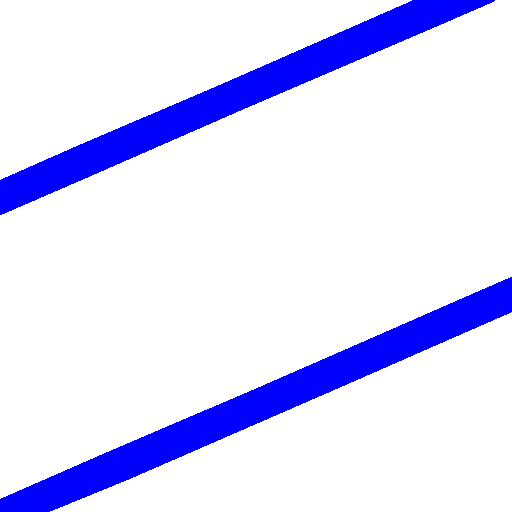} & \includegraphics[width=\linewidth]{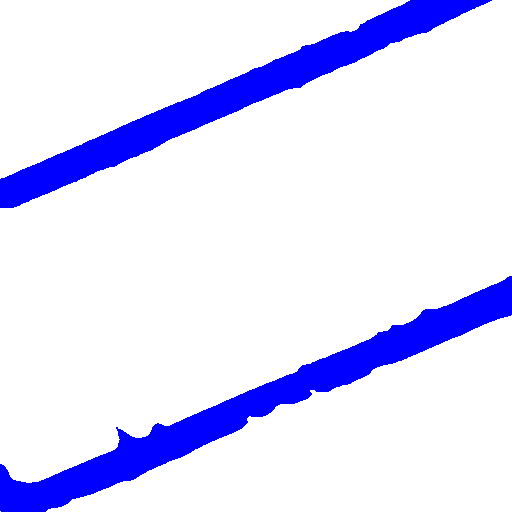} & \includegraphics[width=\linewidth]{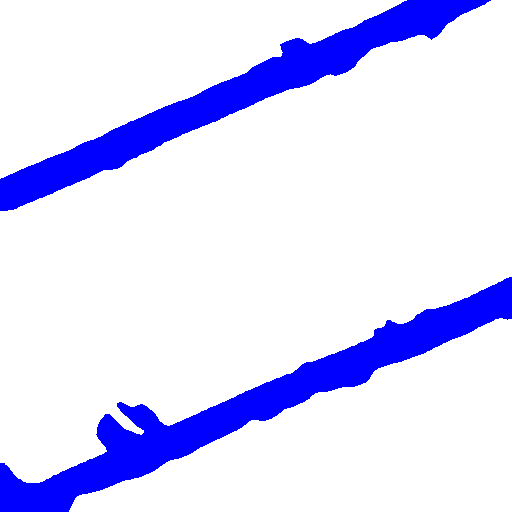} & 
        \includegraphics[width=\linewidth]{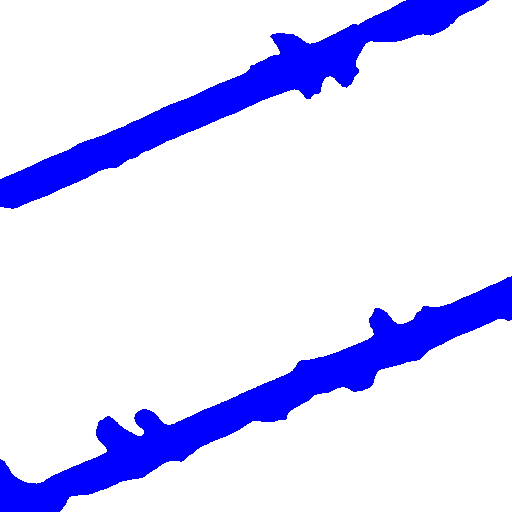} & \includegraphics[width=\linewidth]{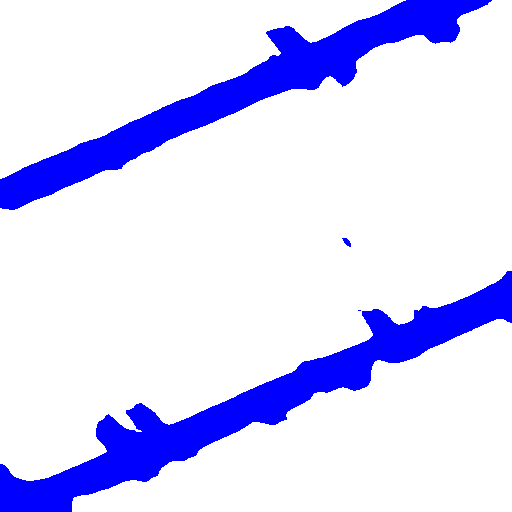} & \includegraphics[width=\linewidth]{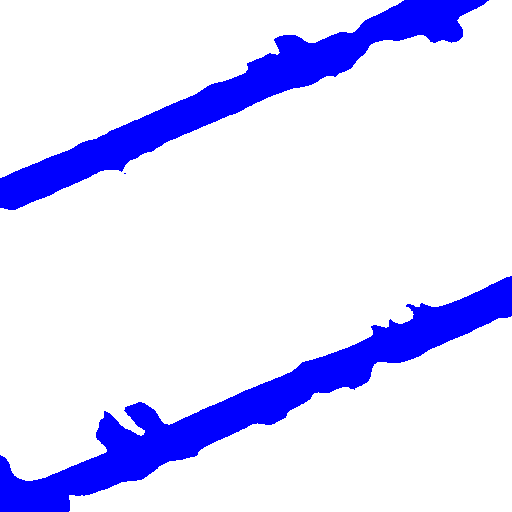} & \includegraphics[width=\linewidth]{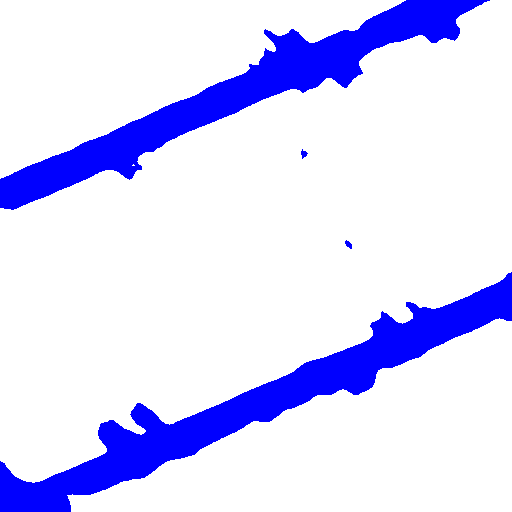} & \includegraphics[width=\linewidth]{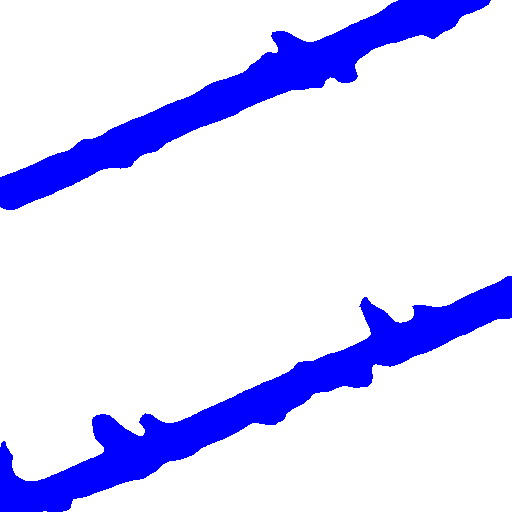}  \\

          &   & \centering	 \textbf{95.68\%} & \centering	 91.94\% & \centering 89.97\% & \centering	 88.74\% & \centering	 89.42\% & \centering	 89.07\% & \centering	 88.36\% & \\

\includegraphics[width=\linewidth]{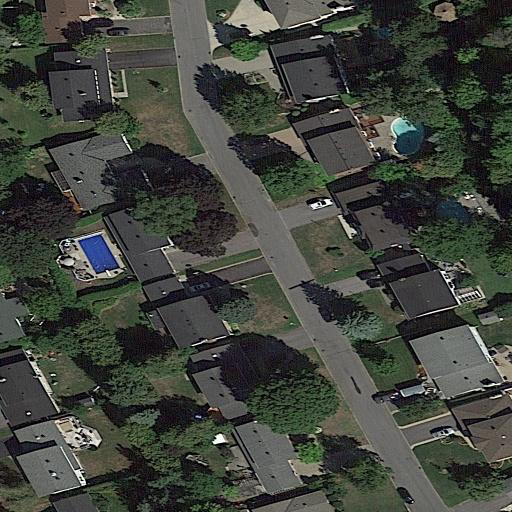} & \includegraphics[width=\linewidth]{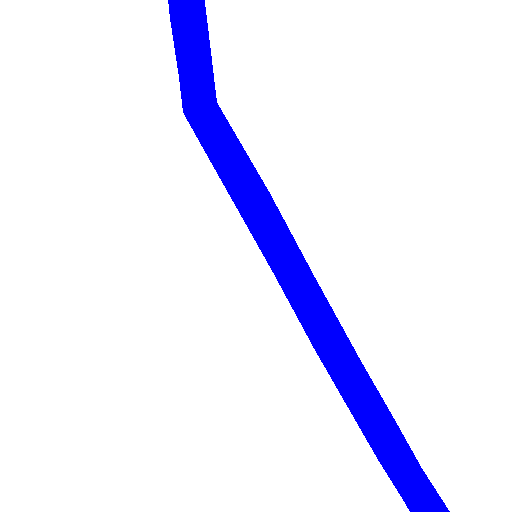} & \includegraphics[width=\linewidth]{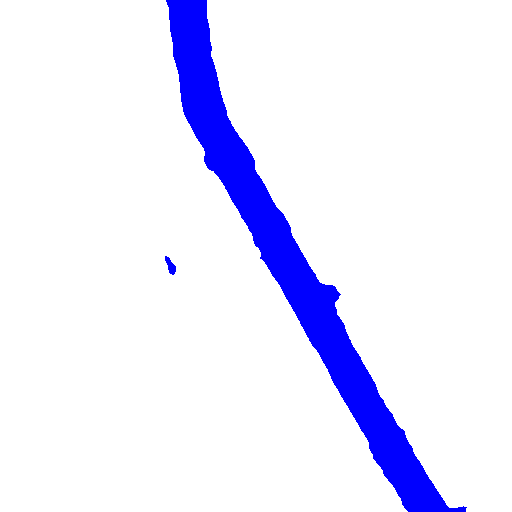} & \includegraphics[width=\linewidth]{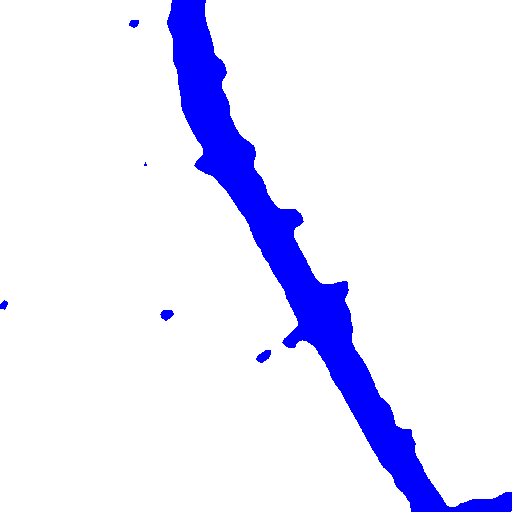} & \includegraphics[width=\linewidth]{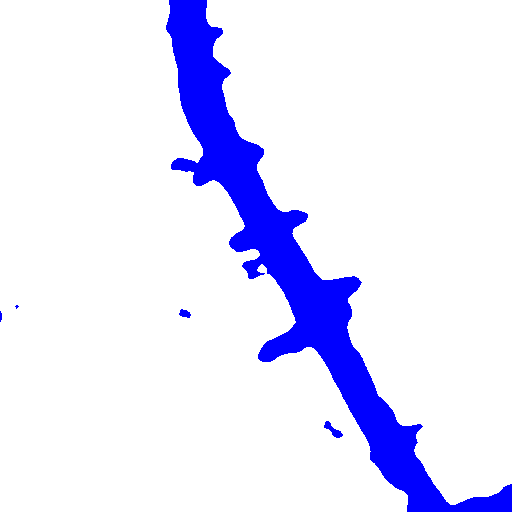} & 
        \includegraphics[width=\linewidth]{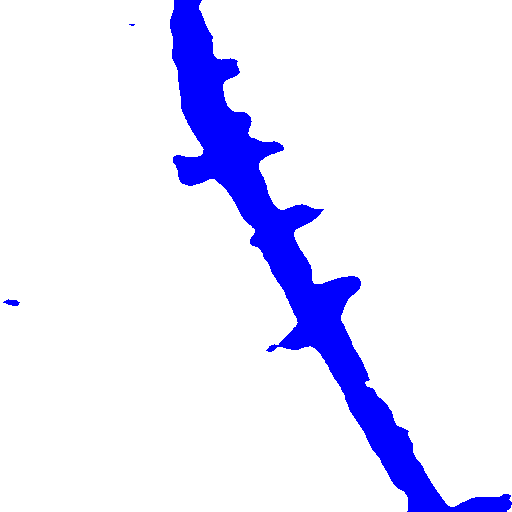} & \includegraphics[width=\linewidth]{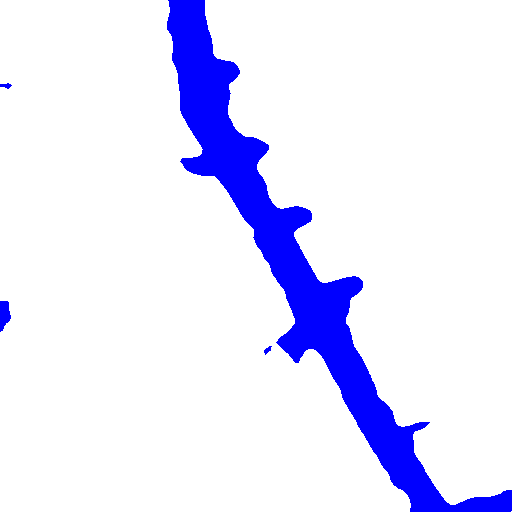} & \includegraphics[width=\linewidth]{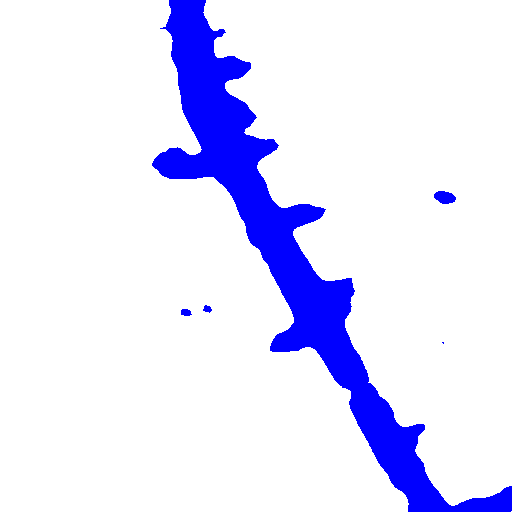} & \includegraphics[width=\linewidth]{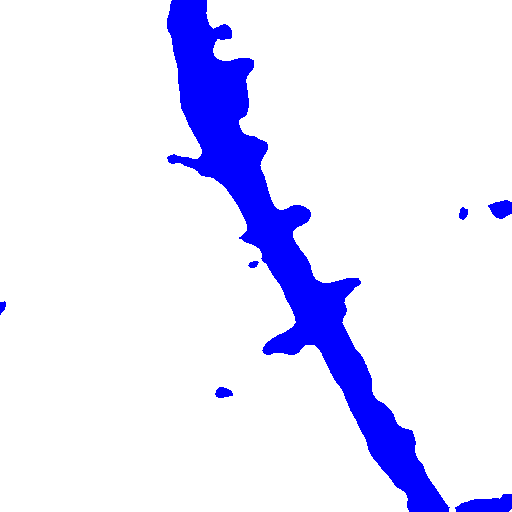} & \\

          &   & \centering	\textbf{95.38\%} & \centering	 88.55\% & \centering  83.74\% & \centering	 84.37\% & \centering	 84.88\% & \centering	 81.29\% & \centering	 82.33\% & \\
          
    \end{tabular} \vspace{-6pt} 
     \caption{Visual comparison between the proposed method and state-of-the-art approaches on the Vaihingen and RoadNet datasets. The mIoU score for each method’s prediction is provided in the corresponding sub-caption.}
    \label{fig:Visual_results}
\end{figure}
\end{landscape}
\clearpage
}

% As illustrated in Figure \ref{Fig:pseudo_label_correction}, our method significantly improves the quality of pseudo-labels through effective correction. The figure presents three representative scenes, each showing the original RGB image, initial pseudo-labels, corrected pseudo-labels, and the corresponding ground truth. Compared to the initial pseudo-labels, the corrected versions exhibit more accurate object boundaries and improved recognition of small or ambiguous objects, such as pedestrians, traffic lights, and poles. Notably, structural consistency is enhanced, especially in challenging regions like occlusions, intersections, and object edges. In several examples, objects that were misclassified or missed entirely in the pseudo-labels are successfully recovered after correction, bringing the predictions much closer to the ground truth. These results qualitatively demonstrate the effectiveness of the proposed framework in refining noisy pseudo-labels and enhancing the reliability of supervision in semi-supervised semantic segmentation tasks.

% \begin{figure*}[ht]
%     \centering
%     \includegraphics[width=\linewidth]{Picture 1.png}
%     \caption{Visualisation of pseudo-label correction. The images, displayed from left to right, represent RGB images, pseudo-labels, corrected pseudo-labels, and ground truth.}
% \label{Fig:pseudo_label_correction}
% \end{figure*}

\section{Ablation Study and Discussion}
\label{Sec:ablation}

We conducted an ablation study on the Vaihingen dataset and provided the results in Table \ref{tab:ablation_study_city} for different combinations of the components of the proposed hybrid method. Prior to applying the proposed active learning process, implementing only the TSF semi-supervised learning method with randomly sampled 15\% of the labelled data, which serves as a practical starting percentage for labelled data \cite{sener2017active, sinha2019variational, seung1992query, xie2020deal}, results in an mIoU of 53.91\%. By incorporating active learning (AL), which identifies erroneous regions using a 0.7 threshold on the predicted probability map (the threshold is consistent with that used in PLAR) and increases the labelled data by 9\% (resulting in a total of 24\%), a modest improvement of 1.66\% in mIoU was achieved, raising it to 55.57\%. The addition of the EMD module further enhanced performance, yielding an additional 0.61\% increase in mIoU and bringing the total to 56.18\%. This modest improvement from integrating EMD indicates that, although EMD provides error locations that are complementary to the confidence-based method, the differences between the error maps produced by the two different approaches are small. Notably, the proposed pseudo-label auto-refinement strategy significantly enhances performance, delivering a 5.82\% mIoU increase, raising the total to 61.39\%, compared to using only TSF with AL. Finally, by integrating all the proposed modules, the hybrid active SSL approach achieves high performance with a mIoU of 61.47\% for the Vaihingen dataset.

\begin{table}
    \centering\setlength{\tabcolsep}{3pt}
    \caption{Effectiveness of the proposed modules by IoU for each class and mIoU (\%). TSF: Teacher-Student-Friend framework; AL: Active Learning; PLAR: Pseudo-Label Auto-Refinement module; EMD: Error Mask Decoder; \checkmark: deploy the module. The best results are highlighted in \textbf{bold}.}
    \begin{tabular}{@{}ccccccccccc@{}}
    \hline
            TSF	       &AL	&PLAR	&EMD	& Road & 
\makecell[c]{Impervious \\ surfaces}& \makecell[c]{Low \\vegetation } & Tree & Car & \makecell[c]{Clutter/ \\ background} &mIoU \\
           \hline \checkmark    &	&	&	& 73.55	& 75.81	& 55.37	& 70.22 &	38.47 &	10.06 &53.91\\
           \hline \checkmark	&\checkmark	&	&	& 74.48&	76.22&	57.26&	70.67&	46&	8.81 &55.57\\
           \hline \checkmark	&\checkmark	&	&\checkmark	& 74.82	&78.11	&58.03	&70.68	&44.87	&10.59 &56.18\\
           \hline \checkmark	&\checkmark	&\checkmark	&	& \textbf{76.27}	&\textbf{82.01}&	59.54&	71.7	&50.84	&\textbf{27.98} &61.39\\
           \hline \checkmark	&\checkmark	&\checkmark	&\checkmark	& 75.95	&81.78	&\textbf{60.4}	&\textbf{72.25}	&\textbf{51.12}	&27.32 &\textbf{61.47}\\
\hline
    \end{tabular} 
    \label{tab:ablation_study_city}
\end{table}

\begin{figure}
\centering
  \begin{tabular}[b]{c}
    \includegraphics[width=.27\linewidth]{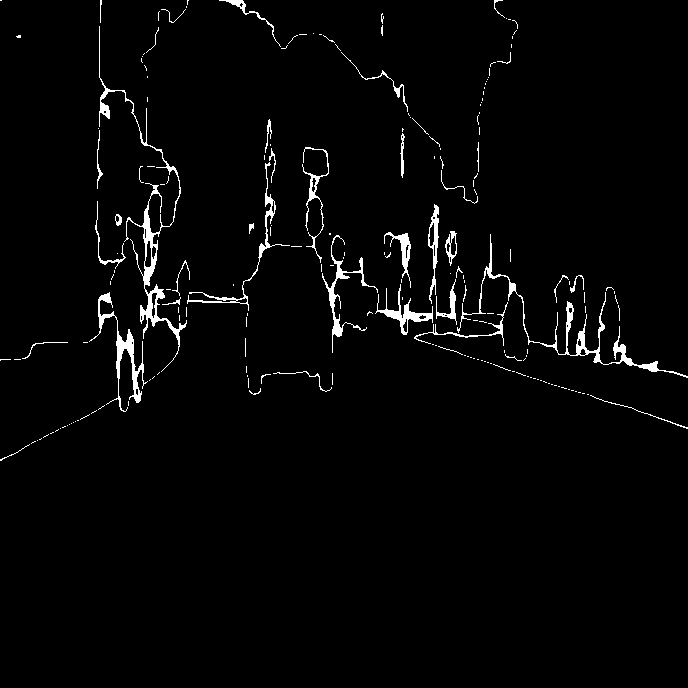} \\
    \small Thres 0.7
  \end{tabular} \hspace{-15pt}
  \begin{tabular}[b]{c}
    \includegraphics[width=.27\linewidth]{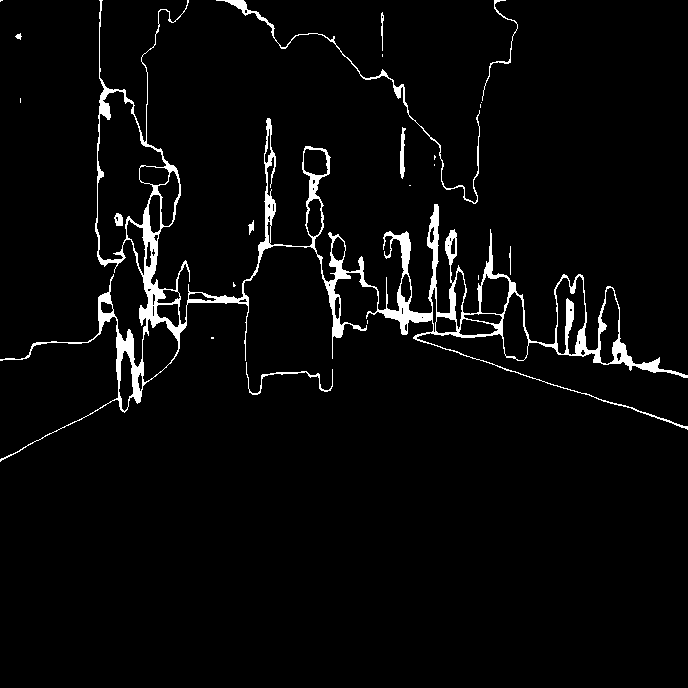} \\
    \small Thres 0.8
  \end{tabular} \hspace{-15pt}
  \begin{tabular}[b]{c}
    \includegraphics[width=.27\linewidth]{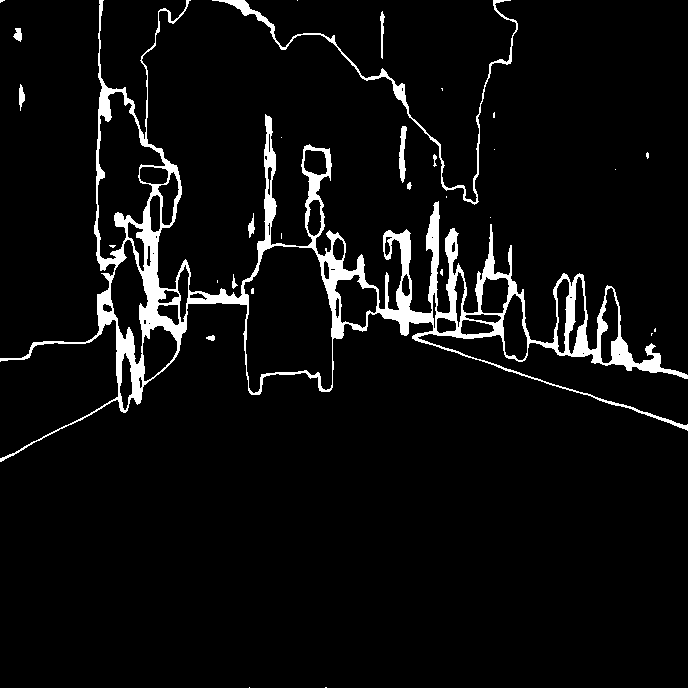} \\
    \small Thres 0.9
  \end{tabular} \hspace{-10pt}
  \begin{tabular}[b]{c}
    \includegraphics[width=.27\linewidth]{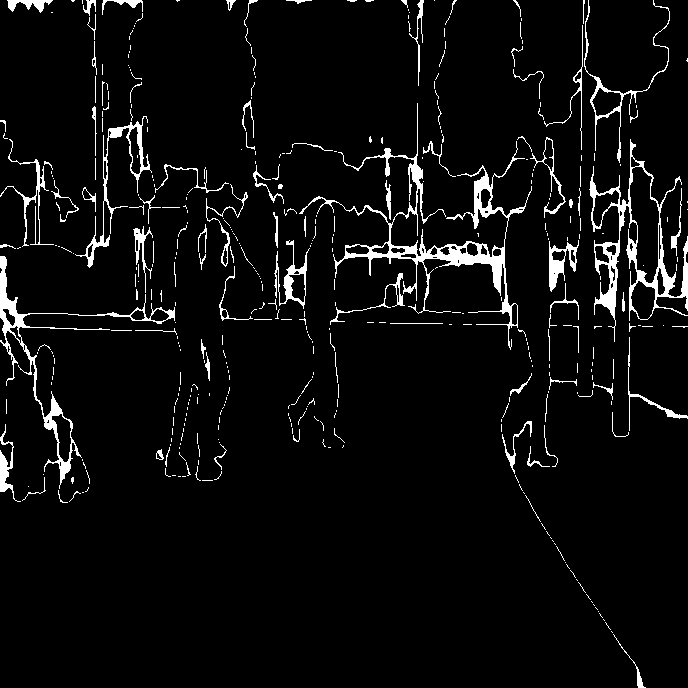} \\
    \small Thres 0.7
  \end{tabular} \hspace{-15pt}
  \begin{tabular}[b]{c}
    \includegraphics[width=.27\linewidth]{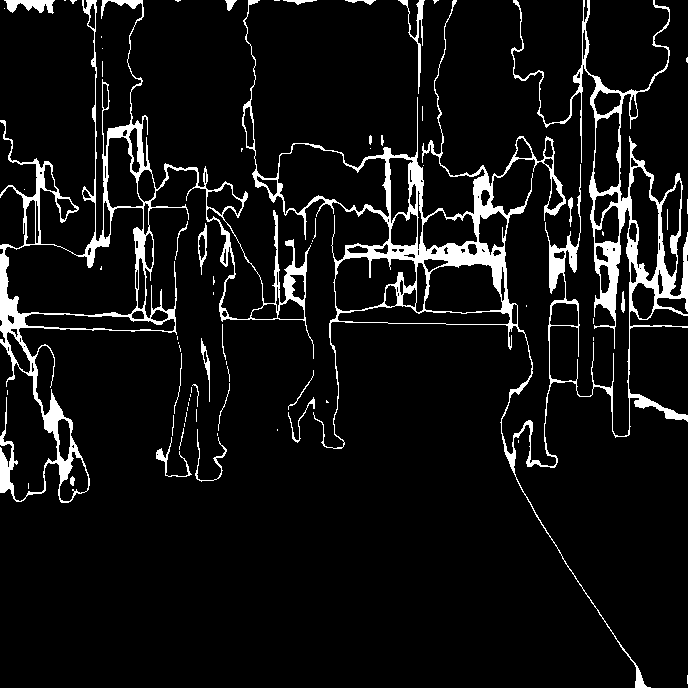} \\
    \small Thres 0.8
  \end{tabular} \hspace{-15pt}
  \begin{tabular}[b]{c}
    \includegraphics[width=.27\linewidth]{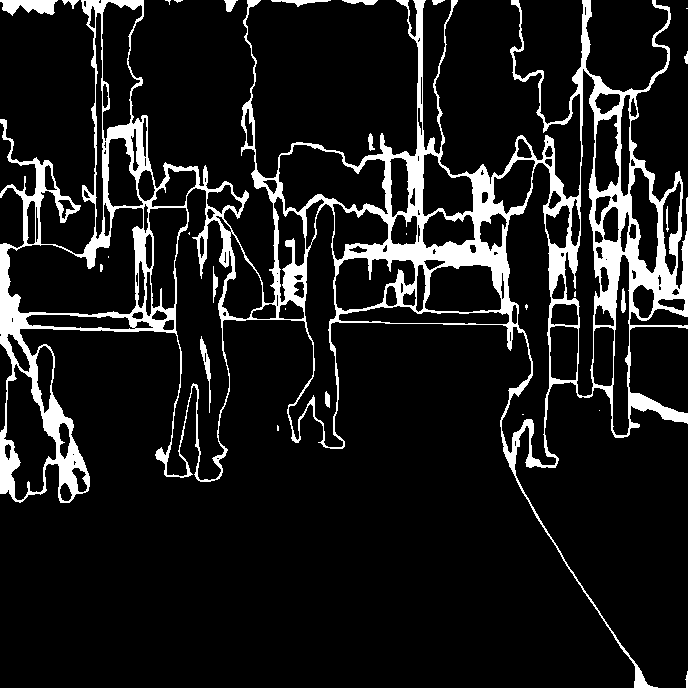} \\
    \small Thres 0.9
  \end{tabular} \hspace{-15pt}
\caption{Probability error maps generated using different thresholds.}
\label{fig:error_mask}
\end{figure}

We present the effectiveness of threshold selection in probability-based error map generation on the CityScapes, Vaihingen and RoadNet datasets. In active learning, threshold selection primarily determines the number of pixels chosen for correction in each pseudo label. On the other hand, we also want to explore how varying the threshold influences performance when the amount of labelled data is fixed. For the Cityscapes dataset, the labelling budget is fixed at 24\% to ensure a fair comparison with benchmark methods, as the state-of-the-art S4AL method \cite{rangnekar2023semantic} also uses 24\% labelled data for this dataset. Therefore, we use Cityscapes to analyse the impact of threshold variation under a fixed labelling ratio. As shown in Table \ref{tab:ablation_study_city_thre}, the observed differences in mIoU across experiments with varying thresholds are relatively small, as they use the same labelled ratio. However, the results indicate that a higher threshold yields slightly better mIoU performance. A larger threshold results in a broader error region being identified in the error map, as shown in Figure \ref{fig:error_mask}. This suggests that exhausting the labelling budget earlier can be beneficial, as more error regions are corrected during the initial training iterations. Consequently, the model benefits from learning with higher-quality pseudo-labels at an early stage, which contributes to improved overall performance. %Notably, the probability-based error map is not applied during the first two epochs.

\begin{table}
    \centering\setlength{\tabcolsep}{1.5pt}
        \caption{Effectiveness of threshold selection in confidence error map generation on the CityScapes validation set with a metric of IoU for each class and mIoU (\%). The best results are highlighted in \textbf{bold}. }
    \label{tab:ablation_study_city_thre}
    \begin{tabular}{@{}ccccccccccc@{}}
\toprule Threshold & Road & 
\makecell[l]{Side \\ walk}& Building & Wall & Fence & Pole & \makecell[l]{Traffic \\ Light} & \makecell[l]{Traffic \\ Sign} & Vegetation & Terrain \\
\toprule 0.5 & 97.20   & 78.06   & 88.91   & 48.57   & 46.32   & 45.14   & 52.42   & 66.35   & 90.16   & 58.67   \\
  0.6 & 97.34   & 78.94   & 88.92   & 45.04   & 48.30   & 45.05   & 52.48   & 66.27   & 90.18   & 59.46   \\
  0.7 & 97.23   & 78.32   & 89.05   & 46.60   & 48.46   & 44.56   & 52.73   & 66.07   & 90.03   & 58.57   \\
  0.8&97.34   & 78.97   & 88.85   & 46.71   & 47.64   & 45.17   & 52.62   & 66.61   & 90.07   & 59.29   \\
  0.9&97.44   & 79.74   & 89.12   & 45.21   & 50.66   & 45.11   & 53.04   & 66.32   & 90.17   & 59.68   \\

\bottomrule \bottomrule
&Sky & 
\makecell[l]{Pedes- \\ trian}  & Rider & Car & Truck & Bus & Train & \makecell[l]{Motor \\ Cycle} & Bicycle & $\mathrm{mIoU}$ \\\toprule 
0.5  & 92.94    & 70.67    & 46.40    & 91.72    & 50.49    & 67.51    & 41.79    & 44.62    & 68.14    & 65.58 \\
0.6 & 92.66    & 70.46    & 45.75    & 91.91    & 52.20    & 68.84    & 46.93    & 42.59    & 67.45    & 65.83 \\
0.7& 93.01    & 70.22    & 45.78    & 91.67    & 53.03    & 70.99    & 48.39    & 40.07    & 67.38    & 65.90 \\

0.8&93.02    & 70.43    & 46.58    & 91.97    & 54.48    & 68.91    & 46.25    & 40.88    & 68.09    & 65.99   \\
0.9&93.01    & 70.41    & 46.42    & 91.78    & 53.58    & 70.53    & 42.99    & 44.97    & 67.94    & 66.22 \\

\bottomrule
\end{tabular}
\end{table}

To further investigate the influence of threshold selection for confidence-based error map generation on the segmentation performance, we conduct an ablation study on the Vaihingen and RoadNet datasets without fixing the labelling budget. In this setup, increasing the threshold allows more error regions of pseudo labels to be corrected, resulting in a larger proportion of labelled data used. The results are summarised in Table \ref{tab:ablation_study_vaihingen} and Table \ref{tab:ablation_study_roadnet}.

Specifically, as shown in Table \ref{tab:ablation_study_vaihingen}, increasing the threshold from 0.5 to 0.9 consistently improves the overall performance. The mIoU rises from 56.14\% to 63.09\%, accompanied by an increase in the labelling proportion from 18\% to 33\%. The class-wise IoU scores demonstrate that the most significant improvements occur in the Tree, Car, and Clutter/background categories—classes that typically contain fine-grained structures and uncertain boundaries. This indicates that correcting a greater number of low-confidence regions enhances the model’s ability to learn challenging categories. Table \ref{tab:ablation_study_roadnet} presents the results for the RoadNet dataset, which contains fewer classes and more homogeneous regions. A similar trend is observed: increasing the threshold yields better mIoU, rising from 79.25\% at a threshold of 0.5 to 88.75\% at 0.9. 

\begin{table}
    \centering\setlength{\tabcolsep}{3pt}
    \caption{Effectiveness of threshold selection in confidence error map generation on Vaihingen dataset by IoU for each class and mIoU (\%). The percentages in the ``Label" column indicate the proportion of labels used in the method relative to the full labels. The best results are highlighted in \textbf{bold}.}
    \begin{tabular}{@{}ccccccccc@{}}
    \hline
         Threshold& Label (\%) & Road & 
\makecell[c]{Impervious \\ surfaces}& \makecell[c]{Low \\vegetation } & Tree & Car & \makecell[c]{Clutter/ \\ background} &mIoU \\
           \hline  0.5& 18\% & 74.79 & 78.37 & 57.43 & 70.67 & 43.88 & 11.70 & 56.14 \\
           \hline 0.6& 21\% & 76.51   & 80.51   & 58.81   & 71.17   & 47.92   & 11.80   & 57.79 \\
           \hline 0.7& 24\% & 75.95 & 81.78 & 60.40 & 72.25 & 51.12 & 27.32 & 61.47 \\
           \hline 0.8& 28\% & 77.20   & 82.56   & 61.07   & 72.64   & 51.15   & 28.53   & 62.19 \\
           \hline  0.9& 33\% & \textbf{77.15}   & \textbf{82.78}   & \textbf{63.07}   & \textbf{74.13}   & \textbf{53.50}   & \textbf{27.89}   & \textbf{63.09} \\
\hline
    \end{tabular} 
    \label{tab:ablation_study_vaihingen}
\end{table}

\begin{table}
    \centering
        \caption{Effectiveness of threshold selection in confidence error map generation on RoadNet dataset by IoU for each class and mIoU (\%). The percentages in the ``Label" column indicate the proportion of labels used in the method relative to the full labels.  The best results are highlighted in \textbf{bold}.}
    \label{tab:ablation_study_roadnet}
    \begin{tabular}{ccccc}
\toprule Threshold & Label (\%)  & Background & Road &  mIoU  \\  
\toprule 0.5& 26\% &94.17 & 64.33 & 79.25\\\hline
 0.6&28\%&96.36 & 76.98 & 86.67 \\\hline
0.7&30\%&96.66 & 78.47 & 87.56 \\\hline
0.8&31\%&96.83 & 79.66 & 88.25 \\\hline
0.9&33\%&\textbf{96.98} & \textbf{80.52} & \textbf{88.75}  \\

\bottomrule
\end{tabular}
\end{table}

\begin{figure}
    \centering
    \includegraphics[width=0.48\linewidth]{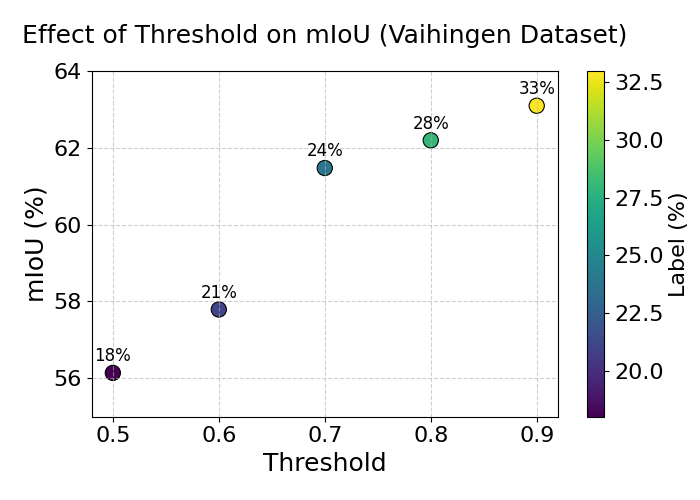}
    \includegraphics[width=0.48\linewidth]{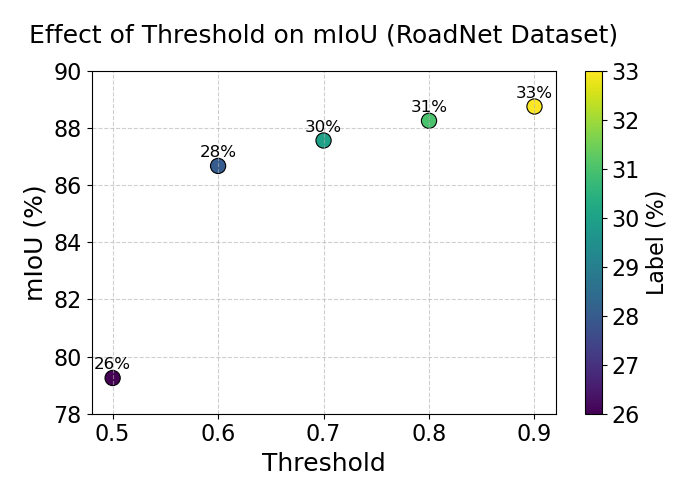}
    \caption{Illustration of how mIoU varies with increasing thresholds and the corresponding proportion of labelled data used for both the RoadNet and Vaihingen datasets.}
    \label{fig:ablation_figures}
\end{figure}

We illustrate how mIoU varies with threshold increments and the associated proportion of labelled data used in our framework for both the RoadNet and Vaihingen datasets in Figure \ref{fig:ablation_figures}. Across both datasets, higher thresholds consistently lead to improved mIoU performance, demonstrating that expanding the corrected label set positively impacts model learning. However, the rate of improvement differs: Vaihingen benefits from a steady gain with increasing thresholds, while RoadNet exhibits rapid improvement at lower thresholds followed by saturation. This difference can be attributed to dataset complexity—complex urban scenes (e.g., Vaihingen) require more corrected samples to capture diverse structures, whereas simpler environments (e.g., RoadNet) reach high accuracy earlier.

\begin{figure}
    \centering
    \includegraphics[width=0.8\linewidth]{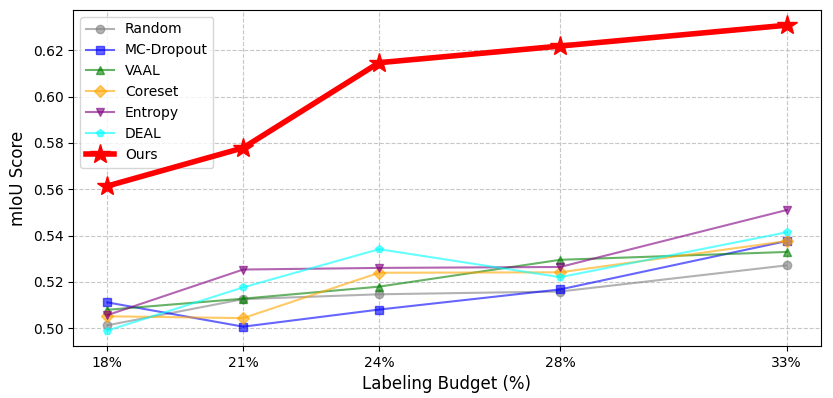}
    \caption{The mIoU performance comparison of various active learning strategies on the Vaihingen dataset across different labelling budget ratios.}
    \label{fig:label_ratio}
\end{figure}

Furthermore, a systemic performance comparison under various labelling budget ratios is shown in Table \ref{tab:vaihingen_final} and figure \ref{fig:label_ratio}. Our proposed method significantly outperforms all compared active learning strategies across all labelling budgets. Remarkably, with only 18\% of labelled data, our method achieves an mIoU of 56.14\%, which already surpasses the performance of all other methods using 33\% of the data. This indicates that our approach can identify extremely informative samples, substantially reducing the annotation effort. At the 33\% budget, our method reaches 63.09\%, a significant gain of 7.99\% over the second-best Entropy-based strategy, proving its superior robustness in complex urban remote sensing scenarios.

\begin{table}[htbp]
\centering
\caption{Performance comparison on the Vaihingen dataset under various labelling budget ratios.}
\label{tab:vaihingen_final}
\begin{tabular}{lccccc}
\toprule
Strategy & 18\% & 21\% & 24\% & 28\% & 33\% \\ 
\midrule
Random         & 50.12\% & 51.25\% & 51.46\% & 51.58\% & 52.71\% \\
Entropy        & 50.56\% & 52.53\% & 52.60\% & 52.64\% & 55.10\% \\
Core-Set \cite{sener2017active}        & 50.51\% & 50.43\% & 52.39\% & 52.41\% & 53.76\% \\
VAAL \cite{sinha2019variational}           & 50.79\% & 51.27\% & 51.79\% & 52.95\% & 53.29\% \\
QBC \cite{seung1992query}            & 51.11\% & 50.06\% & 50.80\% & 51.67\% & 53.76\% \\
DEAL \cite{xie2020deal}           & 49.89\% & 51.76\% & 53.41\% & 52.20\% & 54.14\% \\

\midrule
\textbf{Ours}  & \textbf{56.14\%} & \textbf{57.79\%} & \textbf{61.47\%} & \textbf{62.19\%} & \textbf{63.09\%} \\
\bottomrule
\end{tabular}
\end{table}

\section{Conclusion}
In this paper, we proposed a hybrid methodology that combines an enhanced semi-supervised learning (SSL) framework, TSF, with a novel active learning strategy designed to refine pseudo-labels and identify the most informative data samples for annotation. The approach is particularly well-suited for domains where manual labelling is costly and time-consuming, such as remote sensing. By effectively leveraging both labelled and unlabelled data, our method significantly reduces annotation effort while delivering superior semantic segmentation performance on benchmark datasets across both natural and remote sensing imagery.

Experimental results demonstrate the individual and combined effectiveness of the proposed SSL and active learning strategies, establishing the method's utility in tackling large-scale remote sensing tasks where high-resolution imagery and sparse annotations are common. While our PLAR module exploits feature similarity in latent space through Euclidean and Mahalanobis distances for pseudo-label correction, we acknowledge that the choice of feature representations was not exhaustively studied. Future research will explore advanced feature selection techniques and attention-based mechanisms to further improve pseudo-label refinement. Additionally, we plan to investigate strategies to mitigate PLAR's reduced effectiveness in highly imbalanced class distributions, which is a prevalent challenge in real-world remote sensing applications.


% \bibitem{ams}
% {\it{Mathematics into Type}}, American Mathematical Society. Online available: 

% \bibitem{oxford}
% T.W. Chaundy, P.R. Barrett and C. Batey, {\it{The Printing of Mathematics}}, Oxford University Press. London, 1954.

% \bibitem{lacomp}{\it{The \LaTeX Companion}}, by F. Mittelbach and M. Goossens

% \bibitem{mmt}{\it{More Math into LaTeX}}, by G. Gr\"atzer

% \bibitem{amstyle}{\it{AMS-StyleGuide-online.pdf,}} published by the American Mathematical Society

% \bibitem{Sira3}
% H. Sira-Ramirez. ``On the sliding mode control of nonlinear systems,'' \textit{Systems \& Control Letters}, vol. 19, pp. 303--312, 1992.

% \bibitem{Levant}
% A. Levant. ``Exact differentiation of signals with unbounded higher derivatives,''  in \textit{Proceedings of the 45th IEEE Conference on Decision and Control}, San Diego, California, USA, pp. 5585--5590, 2006.

% \bibitem{Cedric}
% M. Fliess, C. Join, and H. Sira-Ramirez. ``Non-linear estimation is easy,'' \textit{International Journal of Modelling, Identification and Control}, vol. 4, no. 1, pp. 12--27, 2008.

% \bibitem{Ortega}
% R. Ortega, A. Astolfi, G. Bastin, and H. Rodriguez. ``Stabilization of food-chain systems using a port-controlled Hamiltonian description,'' in \textit{Proceedings of the American Control Conference}, Chicago, Illinois, USA, pp. 2245--2249, 2000.

% \end{thebibliography}

\bibliographystyle{cas-model2-names}
\bibliography{example_paper}

% \section*{List of Figure Captions}
% \renewcommand{\cftfigpresnum}{Figure~} % 在编号前加 Figure
\listoffigures
% Biography text here without a photo.

% \begin{IEEEbiography}[{\includegraphics[width=1in,height=1.25in,clip,keepaspectratio]{fig1.png}}]{IEEE Publications Technology Team}
% In this paragraph you can place your educational, professional background and research and other interests.\end{IEEEbiography}

\end{document}